%% file: paper.tex
\documentclass[11pt]{article}

\usepackage[final]{acl}
\usepackage{times}
\usepackage{latexsym}
\usepackage[T1]{fontenc}
\usepackage[utf8]{inputenc}
\usepackage{microtype}
\usepackage{graphicx}
\graphicspath{{figures/out/}}
\usepackage{booktabs}
\usepackage{tabularx}
\usepackage{array}
\usepackage{amsmath}
\usepackage{amssymb}
\usepackage{xcolor}
\usepackage{listings}
\usepackage{enumitem}
\usepackage{tikz}
\usepackage{pgfplots}
\usepackage{multirow}
\usepackage{subcaption}
\usepackage{float}
\usepackage{pifont}
\usepackage{makecell}
\usepackage{url}
\newcommand{\cmark}{\ding{51}}
\newcommand{\xmark}{\ding{55}}
\pgfplotsset{compat=1.18}
\usepgfplotslibrary{groupplots,colormaps,fillbetween}
\usetikzlibrary{shapes.geometric, arrows.meta, positioning, calc, patterns, fit, decorations.pathreplacing}

\definecolor{pastelBlue}{HTML}{B3D4FC}
\definecolor{pastelGreen}{HTML}{B8E6C8}
\definecolor{pastelOrange}{HTML}{FDDCB5}
\definecolor{pastelPurple}{HTML}{D4C4E8}
\definecolor{pastelYellow}{HTML}{FFF3B0}
\definecolor{pastelRed}{HTML}{F5B7B1}
\definecolor{pastelGray}{HTML}{D5D8DC}
\definecolor{pastelTeal}{HTML}{A8E6CF}

\definecolor{richNavy}{HTML}{1B3A5C}
\definecolor{richTeal}{HTML}{2E6B8A}
\definecolor{richOrange}{HTML}{D4832A}
\definecolor{richPurple}{HTML}{8C569C}
\definecolor{richRed}{HTML}{C0392B}
\definecolor{richGray}{HTML}{7F7F7F}
\definecolor{richSteelBlue}{HTML}{4682B4}
\definecolor{richGreen}{HTML}{2E8B57}

\lstdefinelanguage{JS}{
  keywords={const, let, var, await, async, return, if, else, for, of, in, function, true, false, null, new, type, interface, Promise},
  sensitive=true,
  morecomment=[l]{//},
  morecomment=[s]{/*}{*/},
  morestring=[b]",
  morestring=[b]',
  morestring=[b]`,
}

\title{EmailBench: A Benchmark for Evaluating LLM Agents\\on Enterprise Email and Productivity Tasks}
\author{
  \textbf{Mukul Singh} \quad 
  \textbf{Mansi Uniyal} \quad
  \textbf{Devin Devlin} \quad 
  \textbf{Wen Xie} \quad 
  \textbf{Big Thadawasin} \\
  \textbf{Ritam Dutt} \quad 
  \textbf{Vivian Lai} \quad 
  \textbf{Hyeonsu B. Kang} \\
  Microsoft \\
  \texttt{\{singhmukul\}@microsoft.com} \\
}

\begin{document}
\maketitle


\begin{abstract}
Enterprise email agents must combine information retrieval, structured state changes, temporal reasoning, and multi-step coordination.
Recent agent benchmarks include productivity tasks, but few center on typed email workflows in a self-contained environment.
We introduce \textbf{EmailBench}, a benchmark of 206 email and productivity scenarios across 16 task categories.
The benchmark couples a typed email API specification with provider-neutral naming, a deterministic synthetic Enron-inspired corpus, and a scenario suite whose topic selection was informed by aggregate task-intent telemetry from an interactive prototype.
Its hybrid evaluation protocol combines 258 executable static assertions with 211 LLM rubrics.
We evaluate eight LM configurations on a fixed single-user corpus.
The best-performing configuration passes only 33.5\% of scenarios despite 99.7\% of its tool calls completing without an observed API failure, with pass rates varying substantially across task categories.
This gap shows that valid tool execution is not equivalent to task completion.
EmailBench\footnote{We plan to publicly release the benchmark and all supporting artifacts.} provides a self-contained environment for end-to-end email-agent evaluation, with broader tool coverage, multi-persona testing, and repeated-run evaluation as future work areas.
\end{abstract}

\section{Introduction}
\label{sec:intro}

Email agents operate over structured records rather than text alone. A request to schedule a meeting from an email thread may require message retrieval, participant resolution, availability checks, event creation, and a follow-up action. Each operation has a typed payload and later calls can depend on identifiers returned by earlier ones. An agent can therefore issue syntactically valid calls while still selecting the wrong records, violating temporal constraints, or stopping before the workflow is complete.

Existing agent benchmarks evaluate web navigation, desktop interaction, database operations, and repository-level software engineering \citep{liu2024agentbench,zhou2024webarena,xie2024osworld,jimenez2024swebench}.
Productivity-focused benchmarks evaluate workflows across office applications, including email and calendars \citep{wang2024officebench,trivedi2024appworld,wang2025odysseybench,li2026clawsbench}.
Within this landscape, we focus on email-centered workflows that require agents to connect information across mailbox, calendar, contact, and task entities and carry out coordinated state changes.

We introduce \textbf{EmailBench}, a self-contained benchmark for evaluating these workflows through a typed API, a linked synthetic corpus, and complementary state-based and semantic checks.
Our contributions are:

\begin{enumerate}[leftmargin=*,itemsep=1pt,topsep=2pt]
\item \textbf{A self-contained benchmark suite.} A typed email API specification with provider-neutral naming is paired with a deterministic synthetic Enron-inspired corpus and 206 scenarios across 16 task categories.
Aggregate prototype intent frequencies influenced topic selection.
\item \textbf{A hybrid evaluation protocol.} 
We combine executable state and answer checks with scenario-specific LLM rubrics to assess task completion, using explicit score aggregation and critical-check requirements.
\item \textbf{An empirical analysis of email-agent reliability.}
Across eight LM configurations, the highest observed task pass rate is 33.5\%, despite high API-level execution success.
Category-level comparisons and recorded trajectories reveal a gap between executing individual operations and completing the requested workflows.
\end{enumerate}

Together, these contributions advance more realistic email-agent evaluation focused on workflow completion, not just API execution.

\section{Related Work}
\label{sec:related}

Agent evaluation increasingly asks whether models can complete tasks through interaction, rather than merely produce correct standalone answers.
This perspective spans heterogeneous environments \citep{liu2024agentbench}, web-based tasks \citep{zhou2024webarena,yoran2024assistantbench}, desktop interaction \citep{xie2024osworld}, and repository-level software engineering \citep{jimenez2024swebench}.
Across these settings, success depends on connecting intermediate actions to an externally evaluated outcome.
EmailBench adopts this task-level perspective for structured email and productivity workflows.

Tool-use and productivity benchmarks examine complementary aspects of this challenge.
Large API collections emphasize selecting and invoking appropriate tools \citep{qin2024toolllm,patil2023gorilla}, while office-automation and simulated-app environments place those operations within broader workflows \citep{wang2024officebench,trivedi2024appworld}.
This distinction matters for email agents: an executable operation may still target the wrong entity or leave a requested workflow incomplete.

Recent enterprise and productivity benchmarks further emphasize cross-functional coordination, long-horizon execution, and stateful services \citep{vishwakarma2025enterprisebench,wang2025odysseybench,li2026clawsbench}.
Email and calendar tasks already appear in this literature, so their inclusion alone does not distinguish EmailBench.
Instead, these benchmarks provide the closest context for its narrower focus on email-centered workflows and the relationships among mailbox, calendar, contact, and task entities.

Earlier email research provides complementary foundations in email corpora and message-content analysis \citep{klimt2004enron,zhang2019email}.
EmailBench shifts the evaluation target from understanding messages to acting on linked productivity records.
Its contribution lies in combining a typed email-centered API, a fixed synthetic cross-entity corpus, 206 scenarios, and executable checks supplemented by semantic rubrics.
This combination supports examining both individual operations and end-to-end task completion, rather than introducing stateful or semantic evaluation in isolation.

\section{EmailBench}
\label{sec:benchmark-main}

EmailBench consists of an API specification, a deterministic corpus, a scenario suite, and a hybrid grader.
Table~\ref{tab:benchmark-summary} summarizes the artifact and distinguishes the specification from the evaluated adapter.
Detailed inventories appear in Appendix~\ref{sec:design}, and complete scenario examples in Appendix~\ref{app:scenario-examples}.

\begin{table}[t]
\centering
\small
\begin{tabularx}{\columnwidth}{@{}lX@{}}
\toprule
\textbf{Component} & \textbf{Scope} \\
\midrule
API specification & 119 operational methods, 2 evaluation utilities, 36 typed interfaces \\
Evaluated adapter & 46 tools spanning messages, calendar, contacts, todo, folders, filters, settings, and identity \\
Corpus & 65 messages, 15 events, 15 contacts, 10 todo tasks, and linked productivity entities \\
Scenarios & 206 tasks, 16 categories, 8 declared API domains, and three assigned difficulty levels \\
Grading & 258 static assertions and 211 LLM rubrics \\
\bottomrule
\end{tabularx}
\caption{EmailBench at a glance. The empirical study evaluates a subset of the API specification.}
\label{tab:benchmark-summary}
\end{table}

\begin{figure}[t]
\centering
\begin{tikzpicture}[node distance=3.5mm,font=\scriptsize,>=Stealth,
box/.style={draw,rounded corners,align=center,text width=6.2cm,minimum height=6.5mm,fill=richNavy!5}]
\node[box] (task) {Scenario query and fixed persona corpus};
\node[box,below=of task] (agent) {Tool-calling agent with up to 15 iterations};
\node[box,below=of agent] (api) {46-tool adapter and changelog state};
\node[box,below=of api] (grade) {Static assertions and LLM rubrics\\Critical checks determine pass or fail};
\draw[->] (task)--(agent);
\draw[->] (agent)--(api);
\draw[->] (api)--(grade);
\end{tikzpicture}
\caption{Evaluated pipeline. The adapter is narrower than the API specification, and all runs use the same persona corpus.}
\label{fig:main-pipeline}
\end{figure}
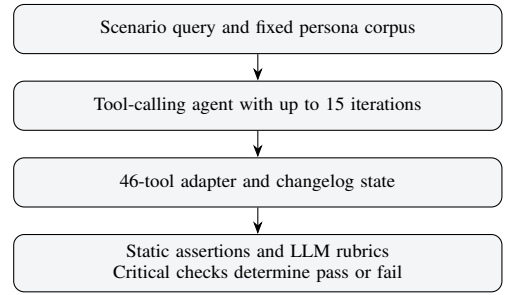

Figure~\ref{fig:main-pipeline} shows the evaluated path from a scenario query through the restricted adapter to the two grading components.

\subsection{API and Corpus}

The \texttt{EmailBenchAPI} specification defines 119 operational methods across ten domain objects plus identity. It also provides \texttt{reset()} and \texttt{snapshot()} for evaluation. The 36 interfaces use provider-neutral names and typed inputs and outputs. Portability across providers is a design goal, not an evaluated result. A snapshot exposes selected entity collections and mutation logs, which support deterministic checks after an agent finishes. The artifact includes an in-memory mutable implementation and a changelog implementation that reads from a frozen corpus and records writes.

The corpus is synthetic and is inspired by themes and organizational relationships in the public Enron email collection \citep{klimt2004enron}. It does not reproduce source emails. The construction process extracts recurring business themes and then generates 65 messages around a fixed reference time. Synthetic calendar events, contacts, todo tasks, boards, filters, and directory records extend the mailbox into a productivity environment. Cross-entity links connect senders to contacts and event attendees, and link project discussions to tasks and meetings. This design supports workflows that require joins across entity types while avoiding the release of private email content.

\paragraph{Evaluation environment.}
The backend pairs a fixed seed corpus with a separate typed action log, preserving agent operations for inspection and grading without modifying the seed data.
Message reads remain tied to the original corpus rather than a fully updated mailbox view. 
Appendix~\ref{app:corpus-details} and Appendix~\ref{app:trajectories} detail the implications for workflows that verify earlier actions through subsequent reads.

\subsection{Scenario Construction}

The benchmark contains 206 scenarios in 16 thematic categories, summarized in Table~\ref{tab:main-categories}.
The initial task inventory was informed by aggregate user task-intent telemetry from a deployed prototype.
Four construction batches added prototype failure cases, underrepresented API domains, additional named personas, and category coverage.
The final suite contains 122 scenarios with one declared domain and 84 with two or more.
Scenario authors assigned 17 tasks to D1, 121 to D2, and 68 to D3.
These labels are metadata rather than experimentally calibrated difficulty levels, and the observed results are not monotonic across them.

\begin{table}[t]
\centering
\small
\begin{tabular}{@{}lr@{\qquad}lr@{}}
\toprule
\textbf{Category} & $n$ & \textbf{Category} & $n$ \\
\midrule
messages-read & 39 & todo & 9 \\
orchestration & 17 & meeting-prep & 9 \\
cross-domain & 15 & inbox-cleanup & 9 \\
messages-write & 15 & folders & 9 \\
multi-domain & 12 & calendar-triage & 8 \\
calendar-read & 12 & settings & 8 \\
calendar-write & 12 & contacts & 12 \\
boards & 10 & filters & 10 \\
\bottomrule
\end{tabular}
\caption{Distribution of the 206 scenarios. Categories denote task themes and are not a pure read-versus-write partition.}
\label{tab:main-categories}
\end{table}

Scenarios span retrieval, aggregation, entity creation, state mutation, and multi-step workflows.
For example, a folder task creates a hierarchy and routes messages using the identifiers returned by folder-creation calls.
A calendar workflow may retrieve participants from messages, identify an available interval, and create an event with nested attendee and time-zone objects.
Appendix~\ref{app:scenario-examples} provides complete task and grading examples.

\subsection{Evaluation Protocol}
\label{sec:grading-main}

Our hybrid evaluation protocol assesses task completion using two complementary sources of evidence: executable static assertions and scenario-specific LLM rubrics.
Static assertions check the final snapshot or textual answer, while LLM rubrics assess correctness, completeness, relevance, format, safety, action completion, and hallucination.
Each scenario may use either component or both.
The suite instantiates 258 static assertions across four of the nine supported types and contains 211 LLM rubrics; 43 scenarios have no static assertions.
Within each component, assertion scores are combined as a weighted mean, with weights defaulting to one.

The static checks comprise 123 answer-substring checks (47.7\%), 101 general predicates over the evaluation snapshot (39.1\%), 32 existence checks (12.4\%), and two checks on sent messages (0.8\%).
These checks verify specified conditions rather than exhaustively establish task completion; for example, confirming that a folder exists does not establish that all intended messages were moved into it.
The 43 scenarios without static checks rely entirely on the LLM component.
Appendix~\ref{app:assertions} reports the complete inventory, including five supported assertion types not instantiated in the evaluated suite.

We combine the two components into a scenario-level score, while critical assertions enforce requirements that cannot be offset by a higher score elsewhere. Let $s_{\mathrm{static}}$ and $s_{\mathrm{LLM}}$ denote the available component scores. When both components are present, the scenario score is
\begin{equation}
s=\tfrac{1}{2}s_{\mathrm{static}}+\tfrac{1}{2}s_{\mathrm{LLM}}.
\end{equation}
When only one component is present, its score is used. A scenario passes when $s\geq0.5$ and every assertion marked critical passes. A critical LLM assertion passes when its score is at least 0.5.

The LLM judge evaluates the query, final answer, and an abbreviated tool trace against each scenario-specific rubric. This complements the executable checks by assessing requirements such as completeness and ordering, although the abbreviated trace limits verification of individual actions. Appendix~\ref{app:judge-protocol} provides the judge prompt, input format, and score-handling rules.

The LLM judge evaluates the query, final answer, and an abbreviated tool trace against scenario-specific rubrics.
Together with executable checks, this provides complementary evidence about whether the agent satisfies the requested workflow, including completeness and ordering requirements.
Appendix~\ref{app:judge-protocol} provides the judge prompt and scoring details, while Appendix~\ref{sec:grader-validation} documents assertion semantics and validation limitations.

\section{Experimental Setup}
\label{sec:setup-main}

We evaluate eight LM configurations under a shared ReAct-style tool-calling protocol \citep{yao2023react}: Claude Sonnet 4.5, Claude Opus 4.1, o3, GPT-5, GPT-5.2, GPT-5.4, GPT-5.4-reasoning, and GPT-5.5-reasoning \citep{anthropic2025sonnet45,anthropic2025opus41,openai2025o3,openai2025gpt5,openai2025gpt52,openai2026gpt54,openai2026gpt55}. Each configuration is evaluated once on all 206 scenarios, using the same agent loop, a maximum of 15 iterations, and a prompt without demonstrations. Agent requests allow up to 4,096 completion tokens. Configuration names follow recorded deployment labels, whose mapping to public model snapshots is not independently verified; Appendix~\ref{sec:setup} documents the identifiers and request settings.

All configurations receive the same 46 JSON-schema tools and are evaluated on a fixed synthetic corpus centered on the ``\emph{Vince Kaminski}'' persona.
The evaluated tool surface is a subset of the broader API specification and excludes boards, directory, and contact-group operations.
Appendix~\ref{app:api-surface} lists the exposed tools, while Appendix~\ref{sec:setup} documents tool-access and persona restrictions.
To examine how these restrictions affect the reported results, Section~\ref{sec:results-main} presents the primary 206-scenario full-suite performance and a post hoc access-screened sensitivity analysis.

Outcomes are scored using the hybrid protocol in Section~\ref{sec:grading-main}.
Static checks inspect the evaluation snapshot and final answer, while the LLM judge assesses scenario-specific rubrics using the answer and abbreviated tool trace.
Judge requests allow up to 2,048 completion tokens.
The judge shares the endpoint used by the configuration labeled GPT-5, leaving possible self-evaluation bias unmeasured.
Appendix~\ref{app:judge-protocol} provides the prompt and configuration details.

Two additional configurations, labeled Phi-4 and Mistral Large, produced no tool invocations and achieved 0\% pass rates in this setup.
We report these additional runs in Appendix~\ref{sec:setup}, but not as part of the eight-configuration comparisons.

\section{Results}
\label{sec:results-main}

\subsection{Overall Performance}

Table~\ref{tab:main-results-compact} reports the complete eight-configuration baseline.
Claude Sonnet 4.5 has the highest observed pass rate on the primary 206-scenario full suite at 33.5\%, followed by Claude Opus 4.1 at 29.6\%.
Rates among the evaluated GPT-5 variants range from 20.9\% to 28.2\%, with no monotonic ordering across all variants. 
We also report a post hoc access-screened sensitivity subset of 152 of the 206 scenarios.
The screen retains a scenario only when every declared domain is exposed by the evaluated adapter, any explicit \texttt{user\_id} identifies the Kaminski persona used by the fixed corpus, and the prompt does not require a known undeclared unavailable contact-group or directory operation.
The three exclusion labels affect 19, 30, and 6 scenarios, respectively; one scenario has overlapping labels, producing a 54-scenario union.
The six undeclared-operation cases are \texttt{OB-H338}, \texttt{OB-X408}, \texttt{OB-X431}, \texttt{OB-X446}, \texttt{OB-X447}, and \texttt{OB-N573}.
Retained-subset rates are lower than removed-slice rates for every configuration (Table~\ref{tab:main-results-compact}).
The removed slice also contains proportionally more LLM-only scenarios (42.6\% versus 13.2\%) and fewer difficulty-1 scenarios (1/54 versus 16/152), showing that screening changes task composition rather than yielding a capability-valid subset.
This descriptive post hoc screen does not repair grading, corpus, or other metadata limitations.
All comparisons are descriptive because each configuration has one run.

\begin{table*}[t]
\centering
\small
\setlength{\tabcolsep}{4pt}
\begin{tabular}{@{}lrrrrrrrrrr@{}}
\toprule
\textbf{Configuration} & \textbf{Pass} & \textbf{Full} & \textbf{Ret.} & \textbf{Rem.} & \textbf{Static} & \textbf{LLM} & \textbf{Calls} & \textbf{Call ok} & \textbf{Clean} & \textbf{Latency} \\
\midrule
Sonnet 4.5 & 69/206 & \textbf{33.5} & 28.3 & 48.1 & .309 & .419 & 5.68 & 99.7 & 94.7 & 27.06 \\
Opus 4.1 & 61/206 & 29.6 & 25.0 & 42.6 & .306 & .419 & 5.37 & 99.7 & 96.6 & 40.56 \\
GPT-5.4-R & 58/206 & 28.2 & 24.3 & 38.9 & .230 & .372 & 3.61 & 96.8 & 88.8 & 21.84 \\
GPT-5.5-R & 57/206 & 27.7 & 25.0 & 35.2 & .300 & .407 & 8.65 & 94.1 & 88.3 & 31.69 \\
GPT-5.4 & 53/206 & 25.7 & 20.4 & 40.7 & .245 & .333 & 2.66 & 95.4 & 89.3 & 17.80 \\
GPT-5.2 & 47/206 & 22.8 & 18.4 & 35.2 & .197 & .286 & 1.86 & 92.4 & 58.7 & 15.84 \\
o3 & 46/206 & 22.3 & 19.7 & 29.6 & .214 & .318 & 6.51 & 95.8 & 79.1 & 67.17 \\
GPT-5 & 43/206 & 20.9 & 18.4 & 27.8 & .204 & .277 & 1.83 & 92.1 & 57.8 & 13.59 \\
\bottomrule
\end{tabular}
\caption{Baseline results from one run per configuration. Pass and Full report the primary 206-scenario suite. Ret. and Rem. are pass percentages on the 152-scenario retained post hoc sensitivity subset and 54-scenario removed slice, respectively. Full, Ret., Rem., call success (Call ok), and clean-execution rate (Clean) are percentages. Static averages the 163 scenarios with static checks, while LLM averages all 206. Calls and latency in seconds are per-scenario means. Clean counts scenarios with at least one call and no observed API failure, using all 206 scenarios as the denominator.}
\label{tab:main-results-compact}
\end{table*}


Beyond pass rates, Table~\ref{tab:main-results-compact} reports static and LLM component scores to distinguish performance on executable checks from rubric-based assessments.
These means cover different scenario sets: static scores average the 163 scenarios with static checks, whereas LLM scores average all 206.
They therefore describe complementary aspects of performance and cannot simply be averaged to recover the mean scenario score.
Appendix~\ref{sec:results} provides the full score breakdown.

The observed ranking summarizes one run per configuration, rather than establishing differences in repeated-run reliability.
Figure~\ref{fig:main-results} in the appendix supplements the point estimates with 95\% Wilson intervals under a scenario-sampling interpretation.
These intervals do not capture model or judge variability; their assumptions and interpretation are detailed in Appendix~\ref{sec:results}.

\subsection{Category and Composition Effects}
Aggregate pass rates conceal substantial variation across task categories (Table~\ref{tab:category-compact}).
Sonnet 4.5 passes 88.9\% of folder tasks (8/9), compared with 16.7\% of calendar-write tasks (2/12), a 72.2-percentage-point difference.
Its strong performance on filters (80.0\%) and message writes (73.3\%) also contrasts with its low orchestration pass rate (17.6\%, 3/17).
These results show that success on some state-changing tasks does not extend uniformly to other write operations or coordinated workflows; a simple read-versus-write distinction misses this variation.
Within the evaluated task mix, these contrasts show that success on some state-changing tasks does not extend uniformly to other write operations or coordinated workflows.
A simple read-versus-write distinction therefore misses substantial variation in performance.

\begin{table}[t]
\centering
\small
\setlength{\tabcolsep}{3pt}
\begin{tabular}{@{}lrrrr@{}}
\toprule
\textbf{Category} & $n$ & \textbf{Son.} & \textbf{5.4-R} & \textbf{GPT-5} \\
\midrule
filters & 10 & 80.0 & 40.0 & 20.0 \\
folders & 9 & 88.9 & 88.9 & 44.4 \\
messages-write & 15 & 73.3 & 46.7 & 53.3 \\
messages-read & 39 & 35.9 & 38.5 & 23.1 \\
calendar-read & 12 & 8.3 & 25.0 & 16.7 \\
calendar-write & 12 & 16.7 & 8.3 & 8.3 \\
orchestration & 17 & 17.6 & 11.8 & 11.8 \\
\bottomrule
\end{tabular}
\caption{Pass rates for representative task categories. The complete eight-configuration table appears in Table~\ref{tab:full-category}.}
\label{tab:category-compact}
\end{table}

Beyond task category, performance also varies with the number of API domains specified in each scenario's metadata (Table~\ref{tab:main-domains}).
For example, a scenario tagged with both messages and calendar has two declared domains, regardless of the agent's actual tool use.
Sonnet's pass rate declines from 37.7\% on one-domain tasks to 34.0\%, 16.7\%, and 14.3\% on tasks declaring two, three, and four domains, respectively.
GPT-5.4-reasoning shows a similar pattern, with corresponding rates of 34.4\%, 26.4\%, 8.3\%, and 0\%.
These results associate broader domain coverage with lower task success, although the cohorts differ in task composition and contain only seven four-domain scenarios.

\begin{table}[t]
\centering
\small
\setlength{\tabcolsep}{4pt}
\begin{tabular}{@{}lrrrr@{}}
\toprule
\textbf{Domains} & \textbf{Son.} & \textbf{Opus} & \textbf{5.4-R} & \textbf{o3} \\
\midrule
1 ($n=122$) & 37.7 & 33.6 & 34.4 & 27.9 \\
2 ($n=53$) & 34.0 & 28.3 & 26.4 & 20.8 \\
3 ($n=24$) & 16.7 & 16.7 & 8.3 & 4.2 \\
4 ($n=7$) & 14.3 & 14.3 & 0.0 & 0.0 \\
\bottomrule
\end{tabular}
\caption{Pass rates by declared domain count. Cohorts differ in task composition and tool availability.}
\label{tab:main-domains}
\end{table}

Unlike domain count, author-assigned difficulty does not show a monotonic relationship with pass rates.
The three tiers designate basic (D1), intermediate (D2), and complex (D3) tasks, but both Sonnet and GPT-5.4-reasoning perform best on D2 and worst on D1 (Table~\ref{tab:difficulty-results}).
Sonnet passes 11.8\%, 42.1\%, and 23.5\% of D1, D2, and D3 tasks, respectively; GPT-5.4-reasoning passes 5.9\%, 37.2\%, and 17.6\%.
The tiers contain 17, 121, and 68 scenarios, respectively.
The present analysis does not establish why D1 tasks have lower pass rates than D2; explaining this pattern requires examining task composition and grading outcomes within each tier.

\subsection{Execution is not task completion}

To examine whether successful tool execution translates into task completion, we compare API-level call outcomes with scenario-level grades.
Sonnet issues 1,170 calls, of which 99.7\% complete without an observed API failure.
Opus also reaches 99.7\%, while the remaining configurations range from 92.1\% to 96.8\%.
Yet scenario pass rates remain between 20.9\% and 33.5\%.
This contrast highlights the distinction between an operation executing successfully and satisfying the user's request: a call can retrieve the wrong entities, apply an incomplete filter, or perform only part of a required workflow.

Recorded trajectories illustrate how this gap arises in practice.
In a successful folder-routing scenario, the agent creates a parent folder and three children, searches by project keyword, and uses the returned folder identifiers to route messages.
In an unsuccessful inbox-triage scenario, all six static checks pass, but the agent reaches the iteration limit before completing the ordered workflow or reporting final counts.
The critical LLM rubric fails, producing a failing outcome despite an aggregate score of 0.50.
Together, these examples illustrate why neither successful API execution nor satisfaction of individual static checks alone establishes complete workflow execution.

\begin{table}[t]
\centering
\small
\begin{tabularx}{\columnwidth}{@{}p{1.4cm}>{\raggedright\arraybackslash}X@{}}
\toprule
\textbf{Scenario} & \textbf{Required workflow and available checks} \\
\midrule
OB-H312 & Aggregate unread percentages by sender; retain rates above 50\% and sort descending. Two static assertions and one rubric check the answer, with a known literal-pattern limitation. \\
\addlinespace
OB-N520 & Create Projects with Raptor, Dynegy, and FERC subfolders, then route keyword matches using returned IDs. Five static assertions and one rubric check the hierarchy and routing. \\
\addlinespace
OB-H307 & Route automated, high-importance, then remaining unread messages in order; exclude Sent Items. Six static assertions and one rubric do not provide equivalent coverage of the full workflow. \\
\bottomrule
\end{tabularx}
\caption{Concrete task and grading requirements. Complete queries and assertions appear in Appendix~\ref{app:scenario-examples}. These cases illustrate coverage differences rather than a sampled error taxonomy.}
\label{tab:main-workflows}
\end{table}

The recorded runs for the tasks in Table~\ref{tab:main-workflows} contrast completed routing with continued activity that does not finish the requested workflow.
In OB-N520, Sonnet makes ten calls over five iterations: one parent-folder creation, three child-folder creations, three searches, and three batch moves, followed by a final report.
Both the static and LLM components score 1.0.
In contrast, OB-H307 uses 29 calls across the full 15 iterations without completing the requested triage.
After initial routing, the agent resumes message listing, creates a second set of the three named folders, and continues retrieval until the limit.
Its final answer announces further retrieval rather than reporting completion.
This case illustrates that additional tool activity need not yield task completion, but does not isolate agent planning from environment constraints such as frozen message reads and folder counters that do not reflect mutations.
Appendix~\ref{app:trajectories} provides the iteration-level evidence and grading rationale.

The reported execution rates measure whether calls complete without an observed API failure, not whether they perform the intended action or satisfy the task.
We classify these outcomes from structured API responses; Appendix~\ref{sec:tool-success} details the classification rules, handling of truncated results, and error-continuation statistics.
Tool engagement, answer production, and task completion must also be distinguished: an agent may answer after an API error, while a recorded pass need not involve a tool call.
Appendix~\ref{sec:graduated} and Appendix~\ref{sec:score-dist} report these indicators and the score distributions.
We next examine engagement and completion timing to distinguish responses without tool calls from attempted but unfinished workflows.

\subsection{Engagement and Completion Timing}
\label{sec:timing-main}

Low task success can coincide with either limited tool engagement or unsuccessful execution after tools are invoked.
To distinguish these patterns, Table~\ref{tab:main-trajectories} separates failures with and without recorded calls and reports when passing trajectories finish.
GPT-5 and GPT-5.2 each have 71 zero-tool failures (34.5\% of all scenarios), compared with seven for Sonnet and four for Opus.
This contrast helps explain why mean call counts alone are insufficient to characterize agent behavior: low averages may reflect responses without tool use rather than efficient workflow execution.
However, the absence of calls identifies behavior, not intent; distinguishing refusals, reports of missing capabilities, and unsupported answers requires inspecting the responses.

\begin{table}[t]
\centering
\small
\setlength{\tabcolsep}{3pt}
\begin{tabular}{@{}lrrr@{}}
\toprule
\textbf{Configuration} & \textbf{Fail/calls} & \textbf{Fail/none} & \textbf{Pass by 5} \\
\midrule
Sonnet 4.5 & 130 & 7 & 62/69 \\
Opus 4.1 & 141 & 4 & 45/61 \\
GPT-5.4-R & 138 & 10 & 47/58 \\
GPT-5.5-R & 144 & 5 & 38/57 \\
GPT-5.4 & 140 & 13 & 48/53 \\
GPT-5.2 & 88 & 71 & 42/47 \\
o3 & 136 & 24 & 30/46 \\
GPT-5 & 92 & 71 & 40/43 \\
\bottomrule
\end{tabular}
\caption{Failures with and without recorded calls, and eventual passes completed within five iterations. Each row's two failure counts plus its total passes sum to 206. The final column conditions on the original run's passing trajectories; it is not a five-iteration rerun.}
\label{tab:main-trajectories}
\end{table}

\begin{figure}[t]
\centering
\includegraphics[width=\columnwidth]{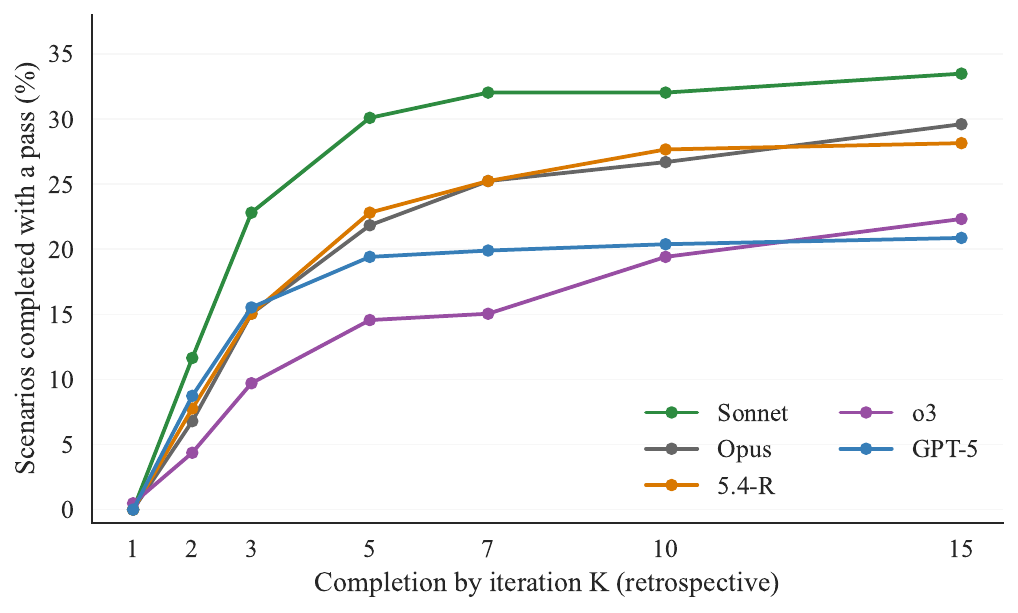}
\caption{Retrospective completion of passing trajectories. Values use all 206 scenarios as the denominator. By iteration five, 65--93\% of each configuration's eventual passes have finished. These curves do not estimate performance under a reduced iteration budget.}
\label{fig:iter-curve}
\end{figure}



Among trajectories that eventually pass, completion timing shows how success is distributed across the interaction budget.
Figure~\ref{fig:iter-curve} counts recorded passing trajectories that finish by iteration $K$ under the original 15-iteration budget.
By iteration five, Sonnet has completed 62 of its 69 passes (89.9\%), compared with 30 of 46 for o3 (65.2\%) and 40 of 43 for GPT-5 (93.0\%).
Late completions therefore account for a larger share of o3's recorded successes than those of Sonnet or GPT-5.
These retrospective curves do not estimate performance under a reduced budget, which could alter agent behavior and final answers.
Appendix~\ref{sec:iteration} provides the complete thresholds and activity profiles.

Tool-call counts provide a complementary view of activity in successful and unsuccessful runs.
Sonnet averages 5.28 calls on passing scenarios and 5.88 on failures; o3 averages 4.63 and 7.06, respectively.
GPT-5 shows the reverse pattern, with 3.02 calls on passes and 1.52 on failures, partly because its many zero-tool failures lower the failure-group mean.
Thus, greater tool activity does not consistently distinguish passing from failing trajectories.
Because these groups contain different tasks, the comparisons do not establish whether additional calls improve task success.
Appendix~\ref{sec:pass-fail} reports the full outcome-conditioned statistics, while Appendix~\ref{sec:recovery} separately examines repeated-call patterns.

\subsection{Shared \& Configuration-Specific Outcomes}

Beyond differences in tool use and completion timing, we examine whether configurations succeed on the same scenarios.
For the selected subset of Sonnet, Opus, GPT-5.4-reasoning, and o3, 24 of 206 scenarios pass for all four, 117 pass for none, and 65 have differing outcomes.
Together, their recorded pass sets cover 89 scenarios (43.2\%), including 20 beyond Sonnet's 69 passes.
This broader coverage indicates complementary successes within the subset, but is not an evaluated ensemble or routing result.

Pairwise comparisons show that this overlap varies across configurations.
Jaccard similarity, defined as the intersection of two pass sets divided by their union, is 0.711 for Sonnet and Opus, compared with 0.530 for Sonnet and GPT-5.4-reasoning and 0.420 for Sonnet and o3.
Thus, Sonnet and Opus share a larger proportion of their successful scenarios than either of these other pairings.
Table~\ref{tab:agreement} reports all six comparisons within the selected subset, which is not the top four configurations by pass rate.

The 117 scenarios failed by all four provide a complementary view of shared weaknesses.
Calendar-domain tasks account for 45.3\% of this set, compared with 34.5\% of the full suite.
Tasks declaring multiple domains account for 46.2\% versus 40.8\%, and D3 tasks account for 40.2\% versus 33.0\%.
Calendar, multi-domain, and D3 tasks are therefore overrepresented among these shared failures.
These overlapping patterns identify candidates for closer inspection, but do not separate agent limitations from tool-access, persona, or grading constraints.
Appendix~\ref{app:hard-scenarios} and Table~\ref{tab:hard-profile} provide the complete cohort profile.

\subsection{Latency and Task Success}

\begin{figure}[t]
\centering
\includegraphics[width=\columnwidth]{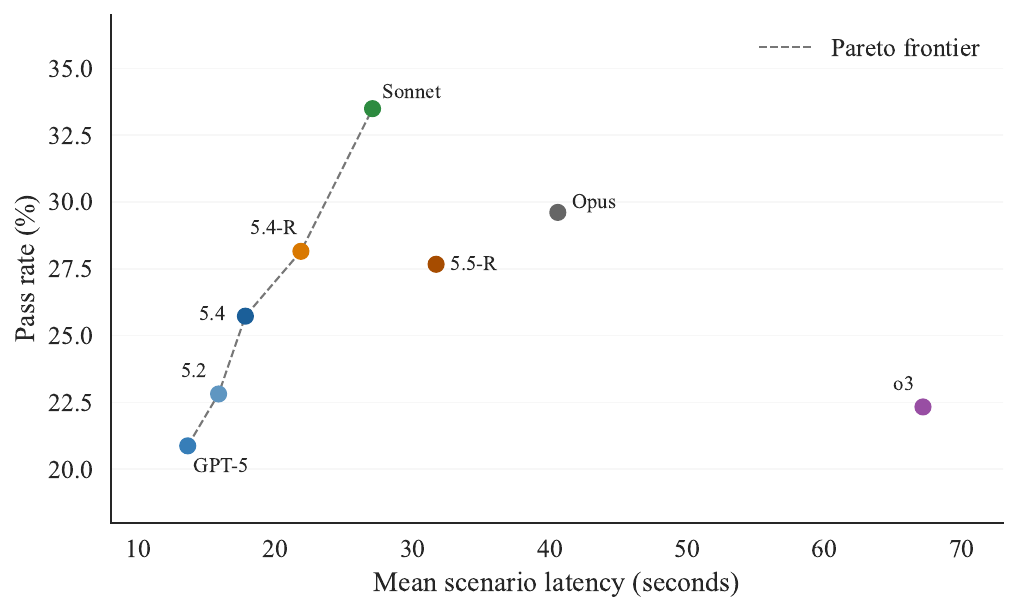}
\caption{Observed tradeoff between mean scenario latency and pass rate among the evaluated configurations.
Sonnet achieves the highest pass rate, while several GPT-5 configurations offer shorter mean latency at lower task success.}
\label{fig:efficiency}
\end{figure}

Finally, we examine task success alongside response time to characterize the practical tradeoffs among configurations.
Sonnet achieves the highest observed pass rate, 33.5\%, at a mean latency of 27.06 seconds per scenario (Figure~\ref{fig:efficiency}).
GPT-5.4-reasoning reaches 28.2\% at 21.84 seconds, while GPT-5 reaches 20.9\% at 13.59 seconds.
Sonnet is both more successful and faster on average than Opus and o3, but several GPT-5 configurations finish sooner at lower pass rates.
The preferred configuration therefore depends on the relative importance of task success and response time.

Recorded latencies include hosted-service effects and task-dependent trajectory lengths, so they characterize the evaluated agent systems rather than isolated model-inference speed.
Table~\ref{tab:tokens} provides the complete latency summary, and Appendix~\ref{sec:efficiency-detail} discusses alternative efficiency measures.

\section{Discussion}
\label{sec:discussion-main}


The central finding is a gap between executing operations and completing the user's workflow.
For Sonnet, 94.7\% of scenarios contain at least one tool call and no observed API failure, yet only 33.5\% pass the task-level grading protocol.
The inbox-triage example illustrates a related distinction: all six static checks pass, but the critical completeness rubric fails.
These results show that successful calls and satisfied individual checks can coexist with an unfinished task.
In enterprise email workflows, the relevant outcome includes selecting the intended entities, preserving dependencies between actions, and completing the requested sequence.

This distinction motivates evaluating agent behavior at multiple levels without treating those levels as interchangeable.
Call-level checks identify execution failures, while targeted state and answer checks verify specified requirements.
Workflow-level assessment must additionally consider whether those requirements collectively cover the user's request.
EmailBench combines these forms of evidence through its hybrid protocol, but the combination does not itself establish agreement with human judgment.
Expert calibration and targeted grader comparisons are needed to assess whether the implemented checks adequately capture task completion.

The findings also motivate interventions that track progress toward the requested outcome, rather than merely increasing tool activity.
Explicit dependency tracking could preserve returned identifiers and action ordering, while completion checks could identify unmet requirements before an agent stops.
These are hypotheses for future experiments, not mechanisms validated by the present results.
Such experiments should provide task-appropriate tools and persona data, ensure that subsequent reads reflect the intended state semantics, and use repeated runs.
Empirical calibration of difficulty labels would also help distinguish anticipated complexity from observed performance.

\subsection{Reproducibility and Availability}

The artifact includes the benchmark implementation, synthetic seed corpus, scenario definitions, executable assertions, LLM rubrics, and evaluation adapter.
The TypeScript environment operates without a live email provider, while model and judge inference requires external services and authorized credentials.
Reported aggregates are derived from recorded outcomes joined to scenario identifiers, supporting inspection of both results and their underlying traces.
The records retain endpoint identifiers and timestamps, but not a complete source-commit and tool-schema manifest.
Appendix~\ref{sec:reproducibility} and Appendix~\ref{app:reproducibility} document the available artifacts, execution requirements, and provenance limits.
A repository-wide license covering all artifacts has not been identified, so unrestricted reuse or redistribution should not be assumed.

\section{Conclusion}
\label{sec:conclusion-main}

We presented EmailBench, a self-contained benchmark for enterprise email and productivity agents.
Across eight configurations in the evaluated single-user setting, the highest observed task pass rate is 33.5\%, despite high API-level execution success.
Category-level results and recorded trajectories demonstrate that executing individual operations does not establish completion of the requested workflow.
EmailBench provides a concrete setting for studying this gap and developing evaluations that connect local execution evidence to end-to-end email-task outcomes.
These findings motivate a task-centered standard for enterprise email agents: success means completing the user's requested workflow, not merely executing its individual operations.

\clearpage
\section*{Limitations} 




The simulated environment isolates email workflows from operational conditions such as network failures, rate limits, eventual consistency, and authentication.
Its 65-message corpus supports controlled evaluation but does not reproduce the retrieval and pagination demands of production mailboxes.
All content and queries are in English, and tasks involve a single user request rather than interactive clarification or correction.

The evaluated setup does not support every scenario requirement.
All runs use the same persona corpus, including 30 scenarios specifying another user, and the supplied tools omit board, directory, and contact-group operations.
Message collection reads remain tied to the seed corpus rather than consistently reflecting recorded mutations, which can affect workflows that verify earlier actions through subsequent reads.
Calendar construction also truncates fractional hours, creating discrepancies with some duration-based rubrics.
Outcomes on affected tasks therefore combine agent behavior with tool-access, corpus, and environment constraints.
The access-screened analysis partially controls for known access and persona mismatches, but remains descriptive and does not resolve other environment or grading issues.

The hybrid grader provides complementary evidence, not an exhaustive or independently validated measure of task correctness.
Static checks verify selected conditions, and 43 scenarios rely entirely on LLM rubrics.
The judge receives abbreviated tool traces, while some answer checks use literal substring matching despite regex-like patterns.
These choices can limit verification or penalize answers for reasons other than task correctness.
We report no independent expert calibration or alternate-judge evaluation.
The judge also shares an endpoint with the configuration labeled GPT-5, leaving possible self-evaluation bias unmeasured.

Each configuration has one recorded run per scenario, so the study does not estimate variability across repeated executions or judgments.
Configuration names follow recorded deployment labels whose mapping to public model snapshots is not independently verified.
Category, domain-count, and difficulty comparisons involve different task compositions and sometimes small cohorts; they do not isolate causal effects.
Likewise, selected trajectories illustrate behaviors rather than establish their prevalence.
The reported results characterize the evaluated agent, environment, and grading protocol together, rather than isolating model behavior from access and grading constraints.

\section*{Ethical Considerations}


The benchmark uses synthetic content inspired by themes and publicly documented organizational relationships in the Enron collection~\citep{klimt2004enron}.
The checked-in corpus contains no original email bodies or private mailbox exports but retains names associated with real people, person-linked email addresses, roles, affiliations, and contact fields.
Synthetic content should therefore not be equated with anonymized data, and generated messages or actions should not be interpreted as statements or conduct of the named individuals.
Public availability of the source collection does not establish unrestricted rights to reuse or redistribute every underlying item or identifier.
The available artifacts contain aggregate prototype intent-frequency counts but do not establish how the underlying telemetry was collected or de-identified; no stronger privacy assurance is supported.

The benchmark executes actions in a simulated environment, but analogous actions in live systems can send messages, move records, or schedule meetings with real consequences.
Its results support studying reliability and evaluation methods, not certifying models for unsupervised deployment.
Adaptation to live systems requires appropriate authorization, confirmation of consequential actions, audit logging, and recovery mechanisms.
Evaluation should also consider whether an agent respects these safeguards, rather than rewarding task completion regardless of how it is achieved.

\newpage
\bibliography{custom}

\clearpage
\appendix

\lstset{
  keywordstyle=\bfseries\color{black},
  commentstyle=\itshape\color{black!60},
  stringstyle=\color{black!75},
  rulecolor=\color{black!35},
  backgroundcolor=\color{black!2}
}

\section{Supplementary Benchmark Design}
\label{sec:design}

This appendix documents the API inventory, corpus construction, scenario metadata, and grading implementation. It also provides the complete evaluated tool surface, prompts, examples, and supplementary analyses needed to interpret the main-paper results.

\subsection{Provider-Neutral API Specification}
\label{sec:api}

The \texttt{EmailBenchAPI} interface defines 119 operational methods across ten domain objects plus identity: 51 reads and 68 writes. Two evaluation utilities, \texttt{reset()} and \texttt{snapshot()}, bring the complete interface to 121 methods. The operational inventory differs from the evaluated 46-tool adapter.
Table~\ref{tab:api-domains} summarizes the coverage, breaking down read and write operations per domain.

\begin{table}[t]
\centering
\small
\begin{tabular}{@{}lrrp{2.3cm}@{}}
\toprule
\textbf{Domain} & \textbf{R} & \textbf{W} & \textbf{Key Ops} \\
\midrule
\texttt{messages}       & 7  & 15 & search, send, reply, flag, batch \\
\texttt{calendar}       & 7  & 11 & create, RSVP, schedule, reminders \\
\texttt{boards}         & 6  & 9 & columns, tasks, assignment \\
\texttt{settings}       & 6  & 9  & vacation, labels, working hrs \\
\texttt{todo}           & 4  & 10  & lists, tasks, checklists \\
\texttt{folders}        & 5  & 5  & create, rename, move, tree \\
\texttt{contacts}       & 5  & 5  & create, folders, search \\
\texttt{filters}        & 2  & 3  & create, update, delete \\
\texttt{directory}      & 5  & 0  & search, manager, reports \\
\texttt{contactGroups}  & 3  & 1  & members, add member \\
\texttt{identity}       & 1  & 0  & current user \\
\midrule
\textbf{Total} & \textbf{51} & \textbf{68} & \textbf{119 methods} \\
\bottomrule
\end{tabular}
\caption{API domain coverage showing read (R) and write (W) method counts.  The 51/68 split excludes the two evaluation utilities. The evaluated adapter is reported separately.}
\label{tab:api-domains}
\end{table}

\paragraph{Design Principles.}
Three principles guide the API design:

\textbf{(1)~Provider neutrality.}
All types use generic terminology, such as \texttt{MailFilter} rather than \texttt{InboxRule}, \texttt{VacationResponder} rather than \texttt{AutomaticReplies}, and \texttt{Label} rather than \texttt{MasterCategory}. A provider-specific implementation can map these contracts to a concrete service.

\textbf{(2)~Typed contracts.}
Every method has fully typed input and output signatures using 36~typed interfaces (\texttt{Message}, \texttt{Event}, \texttt{Contact}, \texttt{TodoTask}, \texttt{BoardTask}, \texttt{MailFolder}, \texttt{MailFilter}, etc.).
Typed contracts prevent ambiguity in both agent code generation and programmatic grading.

\textbf{(3)~Observable state.}
The \texttt{snapshot()} method returns an \texttt{APISnapshot}. It contains selected entity collections and three mutation logs (\texttt{sentMessages}, \texttt{deletedIds}, and \texttt{movedMessages}), which support deterministic post-hoc grading without temporal coupling to agent execution.

\paragraph{Implementations.}
We provide two implementations. \textbf{MockOutlookAPI} (655~LOC) is an in-memory mutable API with OData-style filtering and mutation tracking. \textbf{ChangelogEmailBenchAPI} (1,034~LOC) reads from a frozen corpus and captures writes as typed changelog entries with 66 entry types, providing immutable provenance for grading.

\subsection{Corpus Construction and Adaptation}
\label{sec:corpus}

EmailBench's seed corpus draws on the publicly available Enron email dataset \citep{klimt2004enron} through the following multi-stage process.

\paragraph{Source Material.}
The Enron corpus contains ${\sim}$500,000 emails from 150 employees, released by the Federal Energy Regulatory Commission (FERC) during the Enron investigation.
We use Vince Kaminski, publicly identified as VP of Research, as the primary persona. The synthetic content covers energy trading, risk modeling, and regulatory compliance across messages, meetings, contacts, and tasks. No original or private email body is included, although public names, roles, addresses, and organizational relationships remain.

\paragraph{Adaptation Pipeline.}
The adaptation proceeds in four stages:

\textbf{Stage~1: Theme extraction.}
We analyze the raw Enron maildir to extract recurring themes (energy trading, FERC filings, VaR models, Project Raptor, Dynegy partnership), organizational relationships (Kaminski $\rightarrow$ Skilling, Beck, Kitchen, Lay), and communication patterns (executive summaries, meeting follow-ups, regulatory alerts).

\textbf{Stage~2: Synthetic message generation.}
Rather than directly importing Enron emails, we \emph{synthesize} 65 new messages whose bodies and subjects are fabricated while retaining some publicly documented names, address strings, roles, and organizational relationships.
Messages are generated via a deterministic factory function with controlled parameters: sender, subject template, temporal offset (\texttt{hoursAgo}/\texttt{daysAgo} relative to the reference time 2001-10-15T10:00:00Z), read state (\texttt{isRead = msgSeq \% 3 $\neq$ 0}), importance level, and conversation threading via shared \texttt{conversationId} values.

\textbf{Stage~3: Productivity entity enrichment.}
Since the source corpus contains only emails, we add synthetic calendar events (15), contacts (15), todo tasks (10), board tasks (7), one mail filter, and 15 organizational directory entries. These entity types support productivity workflows that are absent from the source data.
Each synthetic entity is grounded in the message themes: calendar events reference meetings mentioned in emails, contact records correspond to email senders, and todo tasks relate to projects discussed in message threads.

\textbf{Stage~4: Cross-entity interconnection.}
We deliberately introduce cross-references to enable realistic multi-domain scenarios.
Table~\ref{tab:corpus} summarizes the final entity counts and Table~\ref{tab:interconnections} details the interconnection types.

\begin{table}[t]
\centering
\small
\begin{tabular}{@{}lr@{\quad}lr@{}}
\toprule
\textbf{Entity} & \textbf{N} & \textbf{Entity} & \textbf{N} \\
\midrule
Messages       & 65 & Boards         & 2  \\
Events         & 15 & Board columns  & 6  \\
Contacts       & 15 & Board tasks    & 7  \\
Folders        & 6  & Users          & 16 \\
Filters        & 1  & Labels         & 6  \\
Calendars      & 3  & Contact groups & 3  \\
Todo lists     & 3  & Contact folders & 3  \\
Todo tasks     & 10 & Attachments    & 11 \\
\bottomrule
\end{tabular}
\caption{Seed corpus entity counts. Actual message membership is 50 in Inbox and 15 in Sent Items. No message has Archive as its parent folder.}
\label{tab:corpus}
\end{table}

\begin{table}[t]
\centering
\scriptsize
\begin{tabular}{@{}lp{4.8cm}@{}}
\toprule
\textbf{Connection} & \textbf{Description} \\
\midrule
Email $\leftrightarrow$ Calendar & Meeting attendees (Skilling, Beck, Kitchen) also appear as email senders, and threads reference upcoming meetings \\
Email $\leftrightarrow$ Contacts & All major senders have contact records with titles, departments, company \\
Calendar $\leftrightarrow$ Tasks & Todo items reference projects discussed in events (``Raptor'' tasks, ``FERC'' prep) \\
Conflicts & Two events (FERC Review, Dynegy Review) deliberate overlap for conflict detection \\
Mixed signals & System accounts (EnronOnline, GasBank) generate noise requiring differential handling \\
\bottomrule
\end{tabular}
\caption{Cross-entity interconnections enabling multi-domain scenarios.}
\label{tab:interconnections}
\end{table}

\paragraph{Source selection.}
The Enron collection provides a publicly released enterprise setting with documented organizational relationships and multi-topic business communication. EmailBench uses these public themes and relationships as scaffolding for fabricated content. SpamAssassin, holiday-calendar, and conference-schedule data are included as resources for possible extensions but are not part of the reported corpus.

\subsection{Benchmark Construction Methodology}
\label{sec:construction}

EmailBench's 206 scenarios were developed in four batches. Aggregate intent-frequency counts from an Outlook-agent prototype informed the task inventory.

\paragraph{Telemetry-Informed Design.}
Aggregate intent-frequency counts from an Outlook-agent prototype were used to identify common user intents and API call patterns. The available artifacts do not establish how the source telemetry was collected or de-identified.
The top-10 intents by frequency are: (1)~search and summarize messages, (2)~schedule meetings, (3)~forward/reply to threads, (4)~organize inbox into folders, (5)~create and assign tasks, (6)~manage calendar conflicts, (7)~set up mail filters, (8)~update vacation settings, (9)~prepare meeting briefs, and (10)~triage and prioritize incoming mail.
Each of the 16 benchmark categories maps to one or more of these observed intent classes. The aggregate counts informed topic selection but do not establish that the benchmark distribution matches production use.

\paragraph{Four-Batch Construction.}
The scenarios were constructed in four iterative batches:
\textbf{Batch~1} (40~tasks): multi-step operations identified as failure modes during prototype testing.
\textbf{Batch~2} (50~tasks): coverage-gap filling for under-represented domains (calendar writes, contacts, boards).
\textbf{Batch~3} (32~tasks): tasks with multiple named personas. The evaluated runner does not switch corpus identity for these tasks.
\textbf{Batch~4} (84~tasks): systematic gap-filling to ensure $\geq$8 scenarios per category (details in Appendix~\ref{app:gap-fill}).

\paragraph{Available validation evidence.}
The released artifacts include scenario definitions, assertion implementations, and recorded assertion outcomes. They do not include per-scenario review records that would support a claim of independent manual validation. The adapter omissions and assertion limitations described below therefore remain part of the evaluation protocol.

\subsection{Task Taxonomy}
\label{sec:taxonomy}

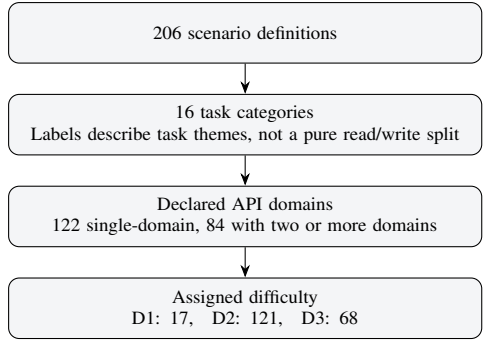
\begin{figure}[t]
\centering
\begin{tikzpicture}[font=\scriptsize,>=Stealth,
box/.style={draw,rounded corners,align=center,text width=6cm,fill=richNavy!5,minimum height=8mm},node distance=4mm]
\node[box] (root) {206 scenario definitions};
\node[box,below=of root] (cat) {16 task categories\\Labels describe task themes, not a pure read/write split};
\node[box,below=of cat] (dom) {Declared API domains\\122 single-domain, 84 with two or more domains};
\node[box,below=of dom] (diff) {Assigned difficulty\\D1: 17,\quad D2: 121,\quad D3: 68};
\draw[->] (root)--(cat);
\draw[->] (cat)--(dom);
\draw[->] (dom)--(diff);
\end{tikzpicture}
\caption{Scenario metadata axes. Category, domain count, and difficulty are separate labels. D3 does not require three or more domains.}
\label{fig:taxonomy}
\end{figure}

We organize the 206 scenarios along three metadata axes (Figure~\ref{fig:taxonomy}): \emph{capability} (what the task tests), \emph{domain} (which API domains are declared), and \emph{assigned difficulty} (the scenario author's qualitative complexity label).

\paragraph{Capability Axis.}
Tasks retrieve and analyze data, create and mutate entities, or combine these actions into workflows. We do not report an aggregate read/write partition without an explicit scenario-level classification. For example, contacts includes mutations, and folders, filters, and settings frequently require writes.

\paragraph{Domain Axis.}
Each scenario is tagged with the API domains it exercises.
The median declared domain count is 1 overall and 2 for D3 scenarios.
The most common co-occurring pairs are messages+calendar (42/206), messages+todo (25/206), and calendar+todo (18/206). A scenario may contribute to several pairs.

\paragraph{Difficulty Axis.}
Table~\ref{tab:difficulty} formalizes the three difficulty tiers.

\begin{table}[t]
\centering
\small
\begin{tabular}{@{}clcc@{}}
\toprule
\textbf{Tier} & \textbf{Definition} & \textbf{N} & \textbf{\%} \\
\midrule
D1 & Basic assigned tasks & 17 & 8 \\
D2 & Intermediate assigned tasks & 121 & 59 \\
D3 & Complex assigned tasks & 68 & 33 \\
\bottomrule
\end{tabular}
\caption{Assigned difficulty tiers from scenario definitions. Overall, 92\% are D2 or D3. Actual tool-call and domain counts need not follow a fixed threshold.}
\label{tab:difficulty}
\end{table}

\subsection{Scenario Design}
\label{sec:scenarios}

EmailBench comprises 206~scenarios across 16~categories (Table~\ref{tab:categories}).

\begin{table}[t]
\centering
\small
\begin{tabular}{@{}lr@{\quad}lr@{}}
\toprule
\textbf{Category} & \textbf{N} & \textbf{Category} & \textbf{N} \\
\midrule
messages-read       & 39 & todo              &  9 \\
orchestration       & 17 & meeting-prep      &  9 \\
cross-domain        & 15 & inbox-cleanup     &  9 \\
messages-write      & 15 & folders           &  9 \\
multi-domain        & 12 & calendar-triage   &  8 \\
calendar-read       & 12 & settings          &  8 \\
calendar-write      & 12 & & \\
contacts            & 12 & & \\
boards              & 10 & & \\
filters             & 10 & & \\
\bottomrule
\end{tabular}
\caption{Scenario category distribution (206 total).}
\label{tab:categories}
\end{table}

\paragraph{Category Design Rationale.}
The categories describe task themes. The following groups are descriptive and may overlap:

\textbf{Read tasks} (\texttt{messages-read}, \texttt{calendar-read}, \texttt{contacts}) test retrieval, filtering, aggregation, and analysis.

\textbf{Write tasks} (\texttt{messages-write}, \texttt{calendar-write}, \texttt{folders}, \texttt{filters}, \texttt{settings}) test state mutations.
Difficulty varies within writes: folder, filter, and settings tasks perform state changes yet achieve higher pass rates than calendar writes. Category labels alone do not define a controlled read-versus-write comparison.

\textbf{Cross-domain tasks} (\texttt{orchestration}, \texttt{cross-domain-analytics}, \texttt{multi-domain-workflow}, \texttt{meeting-prep}, \texttt{inbox-cleanup}, \texttt{calendar-triage}) test whether the agent can correlate data across API domains and execute multi-step pipelines.

\paragraph{Representative Scenario Examples.}
Table~\ref{tab:examples} illustrates three workflows for which recorded evaluation traces are available.

\begin{table}[t]
\centering
\scriptsize
\setlength{\tabcolsep}{3pt}
\begin{tabularx}{\columnwidth}{@{}p{1.05cm}>{\raggedright\arraybackslash}X@{}}
\toprule
ID & Evaluated query and grading \\
\midrule
\textbf{OB-H312} & \textbf{D2, messages-read}. For each person who emailed me, calculate what percentage of their emails to me are unread. Show only people with $>$50\% unread rate, sorted by unread percentage descending.\\
 & 2 static assertions and 1 LLM rubric. \\
\midrule
\textbf{OB-N520} & \textbf{D3, folders}. Create a folder hierarchy: 'Projects' with subfolders 'Raptor', 'Dynegy', and 'FERC'. Then move all emails mentioning each project keyword to the appropriate subfolder.\\
 & 5 static assertions and 1 LLM rubric. \\
\midrule
\textbf{OB-H307} & \textbf{D3, inbox-cleanup}. Set up an email triage system: (1) Create folders 'Action Required', 'FYI Only', and 'Automated'. (2) Move all emails from automated senders (enrononline, legal-tickets, gasbank) to 'Automated'. (3) Move all high-importance emails to 'Action Required'. (4) Move all remaining unread emails to 'FYI Only'. Do NOT move any emails that are already in Sent Items.\\
 & 6 static assertions and 1 LLM rubric. \\
\bottomrule
\end{tabularx}
\caption{Three evaluated scenario definitions with recorded traces. OB-H312 analyzes senders, OB-N520 creates a hierarchy, and OB-H307 performs triage. Appendix~\ref{app:scenario-examples} details their checks.}
\label{tab:examples}
\end{table}

\subsection{Hybrid Grading Pipeline}
\label{sec:grading}

EmailBench evaluates agent performance through two complementary assertion modes, combined into a single score.

\subsubsection{Static Assertions}
The assertion interface supports nine types (Table~\ref{tab:assertions}). The 206 evaluated scenarios instantiate 258 static assertions using four types. Snapshot checks and answer-substring checks have different inputs.

\begin{table}[t]
\centering
\scriptsize
\begin{tabular}{@{}lp{4.6cm}@{}}
\toprule
\textbf{Type} & \textbf{What it checks} \\
\midrule
\texttt{count}           & Collection size matches expected value (eq, gte, lte, gt, lt operators) \\
\texttt{exists}          & Predicate returns true on snapshot \\
\texttt{not-exists}      & Predicate returns false on snapshot \\
\texttt{property}        & Strict equality on a computed value \\
\texttt{sent-message}    & Sent message matches subject/to/body \\
\texttt{moved-message}   & Messages moved to target folder \\
\texttt{deleted}         & Minimum entities deleted \\
\texttt{state-check}     & Arbitrary boolean predicate on snapshot \\
\texttt{return-contains} & Agent answer contains pattern \\
\bottomrule
\end{tabular}
\caption{Nine supported static types. Snapshot checks inspect \texttt{APISnapshot}, while \texttt{return-contains} checks the final answer. Four types occur in the evaluated suite.}
\label{tab:assertions}
\end{table}

\subsubsection{LLM-Judge Assertions}
\label{sec:llm-judge}
For aspects that resist programmatic checking (e.g., ``Generate a professional meeting prep brief''), we use an LLM judge \citep{zheng2023judging} that scores against rubrics across 7 semantic dimensions: \emph{correctness}, \emph{completeness}, \emph{relevance}, \emph{format}, \emph{safety}, \emph{action-taken}, and \emph{no-hallucination}.

The judge receives the scenario query, the agent's text answer, a truncated tool-call trace (arguments capped at 100 characters and results at 200 per call), and the rubric.
It returns a JSON object containing a score in $[0, 1]$ and a one-sentence rationale. The evaluator clamps a parsed score to this range and derives the critical-check outcome by comparing it with the 0.5 threshold. An unparseable response or judge-call error produces a failing zero score. Run logs identify judge deployment \texttt{dev-gpt-5-chat-jj}; the checked-in request omits temperature.

\subsubsection{Scoring}
When both assertion components are present, the final score is
\[
s = \tfrac{1}{2}s_{\text{static}} + \tfrac{1}{2}s_{\text{LLM}}.
\]
When only one component is present, its score is used directly.

A scenario \emph{passes} when $s \geq 0.5$ \textbf{and} all assertions marked \texttt{critical: true} pass.
The critical flag enables hard requirements (e.g., ``the folder must exist'') while allowing soft scores for quality aspects.
The evaluated baseline uses equal component weights when both exist and the available component otherwise. These rules depend on assertion availability, not on whether a task reads or writes. Reported scores and pass flags are taken from the recorded runs.

\subsection{Grading Inventory and Validation Limits}
\label{sec:grader-validation}

Across 206 scenarios, 258 static assertions give a mean of 1.252 per scenario. Forty-three scenarios have no static assertion. The suite contains 211 LLM assertions. Appendix~\ref{app:assertions} gives the static inventory.

Result files include assertion outcomes and LLM-judge rationales. The available artifacts do not include the expert annotations needed to estimate inter-rater agreement or conduct a grader ablation. We therefore make neither claim.

The executed \texttt{return-contains} checker uses literal substring matching, although some scenario patterns are written as regular expressions. For example, OB-H312's percentage pattern is treated literally even when an answer includes a percentage. The reported grades follow these executed semantics. Changing them requires a separately reported regrading analysis. Recorded final answers and patterns permit offline reevaluation of this assertion class.

\section{Supplementary Benchmark Comparison}
\label{sec:comparison}

Table~\ref{tab:comparison} situates EmailBench among general agent, office-workflow, and enterprise benchmarks. The columns report properties described by each benchmark and are not intended as quality rankings.

\begin{table*}[t]
\centering
\scriptsize
\begin{tabular}{@{}lcccccccc@{}}
\toprule
\textbf{Benchmark} & \textbf{Domain} & \textbf{Tasks} & \textbf{API} & \textbf{Typed} & \textbf{Grading} & \textbf{Offline} & \textbf{Multi-D} & \textbf{Corpus} \\
\midrule
AgentBench        & 8 envs         & 1,013 & N/A   & \xmark & Functional & \cmark & \xmark & Varied \\
WebArena          & Web            & 812   & N/A   & \xmark & Match      & \xmark & \xmark & Live sites \\
SWE-bench         & Code           & 2,294 & N/A   & \xmark & Unit tests & \cmark & \xmark & GitHub \\
OfficeBench       & Office apps    & 300   & N/A   & \xmark & Match/exec. & \xmark & \cmark & Synthetic \\
AppWorld          & 9 apps         & 750   & 457   & \cmark & API state  & \cmark & \cmark & Synthetic \\
EnterpriseBench~\citep{vishwakarma2025enterprisebench} & Enterprise & 500 & N/A & N/A & Task-specific & \cmark & \cmark & Synthetic \\
OdysseyBench~\citep{wang2025odysseybench} & Office apps & 602 & N/A & N/A & Task-specific & N/A & \cmark & Mixed \\
ClawsBench~\citep{li2026clawsbench} & 5 services & 44 & 198 & N/A & State/safety & \cmark & \cmark & Synthetic \\
\midrule
\textbf{Ours} & \textbf{Email+} & \textbf{206} & \textbf{119} & \textbf{\cmark} & \textbf{Hybrid} & \textbf{\cmark} & \textbf{\cmark} & \textbf{Adapted} \\
\bottomrule
\end{tabular}
\caption{Comparison with related benchmarks. ``API'' denotes a reported method, endpoint, or route count. Counts are not directly equivalent across abstractions. ``Multi-D'' denotes cross-domain or cross-application scenarios. N/A indicates that the cited paper does not report a directly comparable value.}
\label{tab:comparison}
\end{table*}

The comparison highlights differences in scope and evaluation abstraction. EmailBench centers on email and adjacent productivity domains, while EnterpriseBench covers cross-functional enterprise work, OdysseyBench emphasizes long-horizon office workflows, and ClawsBench evaluates stateful productivity services and safety. These design differences preclude direct comparison of task counts or headline scores.

\section{Supplementary Experimental Results}
\label{sec:experiments}

\subsection{Evaluation Configuration}
\label{sec:setup}

We report eight recorded configurations under labels from three model groups using a standardized ReAct-style tool-calling agent \citep{yao2023react}:
\textbf{OpenAI GPT-5 labels:} GPT-5, GPT-5.2, GPT-5.4, GPT-5.4-reasoning, and GPT-5.5-reasoning \citep{openai2025gpt5,openai2025gpt52,openai2026gpt54,openai2026gpt55}.
\textbf{OpenAI o3 label:} o3 \citep{openai2025o3}.
\textbf{Anthropic labels:} Claude Opus 4.1 and Claude Sonnet 4.5 \citep{anthropic2025opus41,anthropic2025sonnet45}.
The recorded files identify internal gateway endpoints rather than public model snapshots (Table~\ref{tab:run-identifiers}). The saved o3 alias maps to \texttt{dev-gpt-5-reasoning}, so the underlying model family is not independently established. Phi-4 and Mistral Large produced no tool invocations in this configuration and received a 0\% pass rate. They are disclosed here but excluded from the eight-configuration analyses below.

\begin{table*}[t]
\centering
\scriptsize
\begin{tabularx}{\textwidth}{@{}p{2.5cm}p{5.4cm}p{3.2cm}l@{}}
\toprule
\textbf{Configuration label} & \textbf{Internal endpoint ID} & \textbf{Run ID} & \textbf{Start time (UTC)} \\
\midrule
Claude Sonnet 4.5 & \path{dev-anthropic-claude-sonnet-4-5} & \path{run-1777853456125} & 2026-05-04 00:10:56 \\
Claude Opus 4.1 & \path{dev-anthropic-claude-opus-4-1} & \path{run-1777856235806} & 2026-05-04 00:57:15 \\
GPT-5.4-reasoning & \path{dev-gpt-54-reasoning} & \path{run-1777882982599} & 2026-05-04 08:23:02 \\
GPT-5.5-reasoning & \path{dev-gpt-55-reasoning} & \path{run-1777885083982} & 2026-05-04 08:58:03 \\
GPT-5.4 & \path{dev-gpt-54-chat} & \path{run-1777882125342} & 2026-05-04 08:08:45 \\
GPT-5.2 & \path{dev-gpt-52-chat} & \path{run-1777881635032} & 2026-05-04 08:00:35 \\
o3 & \path{dev-gpt-5-reasoning} & \path{run-1777861576649} & 2026-05-04 02:26:16 \\
GPT-5 & \path{dev-gpt-5-chat-jj} & \path{run-1777846519055} & 2026-05-03 22:15:19 \\
\bottomrule
\end{tabularx}
\caption{Identifiers preserved in the recorded result files. Configuration labels reproduce the experiment metadata. The endpoint-to-public-snapshot mapping is not independently verified.}
\label{tab:run-identifiers}
\end{table*}

\paragraph{Agent Architecture.}
The evaluated adapter supplies 46 JSON-schema tools: messages (13), folders (6), filters (4), calendar (7), contacts (4), todo (6), settings (5), and identity (1). It constructs \texttt{ChangelogEmailBenchAPI(CORPUS)} and runs for up to 15 iterations. Board, directory, and contact-group tools are absent.
No few-shot examples are supplied. Every scenario uses the fixed Kaminski corpus: the runner ignores \texttt{scenario.user\_id}. Of 206 scenarios, 34 declare a user ID and 30 specify another user. These results describe this restricted setup, not genuine multi-user evaluation. Expanded tool access or user-specific corpora require a revised experiment and reruns.
Each scenario is run once per model. The checked-in agent request omits temperature and sets \texttt{max\_completion\_tokens=4096}. The 180-second agent bridge limit applies per request, not per scenario. Judge requests use a 120-second subprocess limit and \texttt{max\_completion\_tokens=2048}; the bridge uses a 120-second HTTP timeout.

\subsection{Component Scores and Intervals}
\label{sec:results}

\begin{table}[t]
\centering
\small
\setlength{\tabcolsep}{2.5pt}
\begin{tabular}{@{}lrrrrrr@{}}
\toprule
\textbf{Model} & \textbf{Pass} & \textbf{Rate} & \textbf{Stat} & \textbf{LLM} & \textbf{Scr} & \textbf{Lat} \\
\midrule
Claude Sonnet 4.5  & 69/206 & \textbf{33.5\%} & .309 & .419 & .385 & 27s \\
Claude Opus 4.1    & 61/206 & 29.6\% & .306 & .419 & .383 & 41s \\
GPT-5.4-reasoning  & 58/206 & 28.2\% & .230 & .372 & .336 & 22s \\
GPT-5.5-reasoning  & 57/206 & 27.7\% & .300 & .407 & .374 & 32s \\
GPT-5.4            & 53/206 & 25.7\% & .245 & .333 & .321 & 18s \\
GPT-5.2            & 47/206 & 22.8\% & .197 & .286 & .269 & 16s \\
o3                 & 46/206 & 22.3\% & .214 & .318 & .290 & 67s \\
GPT-5              & 43/206 & 20.9\% & .204 & .277 & .266 & 14s \\
\bottomrule
\end{tabular}
\caption{Recorded results across 206 scenarios. Stat averages the 163 scenarios with static checks, and LLM averages all 206. Scr is the mean scenario score, with equal weights when both components exist and the available component otherwise. Passing requires score $\geq0.5$ and all critical assertions to pass.}
\label{tab:main-results}
\end{table}

\begin{figure}[t]
\centering
\includegraphics[width=\columnwidth]{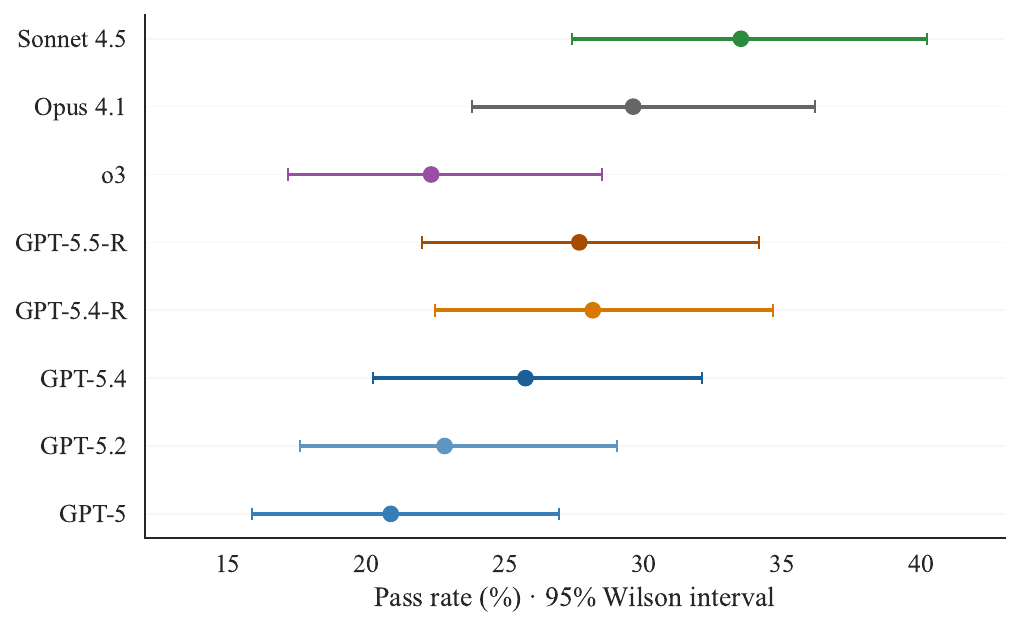}
\caption{Recorded pass rates with two-sided 95\% Wilson score intervals ($n=206$ per model). These intervals describe hypothetical scenario sampling, not repeated-run or judge variability. Sonnet has the highest observed rate, 69/206 (33.5\%).}
\label{fig:main-results}
\end{figure}

Table~\ref{tab:main-results} and Figure~\ref{fig:main-results} present the main results.
For these intervals, let $\hat p=k/n$, $n=206$, and $z=1.959964$. Wilson center and half-width are
\[
\begin{aligned}
c&=\frac{\hat p+z^2/(2n)}{1+z^2/n},\\[3pt]
h&=\frac{z\sqrt{\hat p(1-\hat p)/n+z^2/(4n^2)}}{1+z^2/n}.
\end{aligned}
\]
We report $[c-h,c+h]$ without continuity correction. This binomial interpretation treats scenarios as independent samples. The fixed-suite rate is already known exactly, task dependence is not modeled, and these intervals do not establish pairwise model significance.

Claude Sonnet 4.5 has the highest observed pass rate at 33.5\%, followed by Opus 4.1 at 29.6\%. Across the reported GPT-5 configurations, pass rates range from 20.9\% to 28.2\%. GPT-5.5-reasoning records 8.7 tool calls per scenario on average, compared with 3.6 for GPT-5.4-reasoning. These uncontrolled comparisons do not isolate the effects of model generation, reasoning mode, or tool-call count.

\paragraph{Static and LLM components.}
All configurations show differences between static and LLM component means (Table~\ref{tab:main-results}). Static means are computed only where the component exists, rather than including the runner's default value for scenarios without static checks. Sonnet records a static mean of 0.309 and an LLM mean of 0.419, while GPT-5 records 0.204 and 0.277. These component means do not identify the mechanism of task failure or validate the grader. Critical checks can cause failure even when an aggregate score reaches the threshold.

\subsection{Performance Across Task Categories}
\label{sec:read-write}

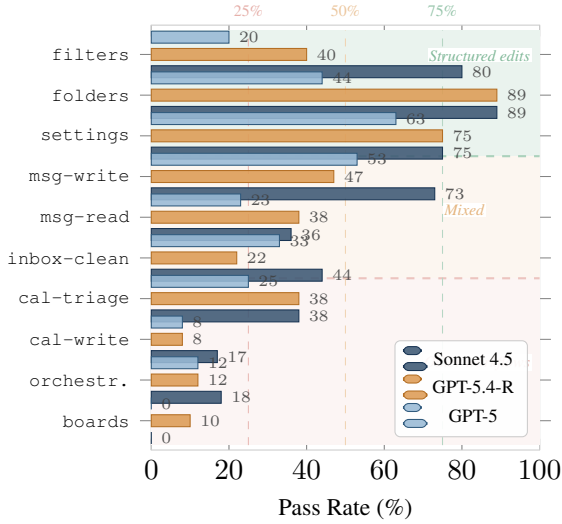
\begin{figure}[t]
\centering
\begin{tikzpicture}
\begin{axis}[
  xbar,
  width=\columnwidth-28pt,
  height=7.2cm,
  bar width=4.5pt,
  xmin=0, xmax=100,
  xlabel={Pass Rate (\%)},
  xlabel style={font=\small},
  ytick=data,
  yticklabels={boards, orchestr., cal-write, cal-triage, inbox-clean, msg-read, msg-write, settings, folders, filters},
  yticklabel style={font=\scriptsize\ttfamily},
  y tick label style={anchor=east},
  legend style={at={(0.97,0.03)}, anchor=south east, font=\scriptsize, draw=black!20, fill=white, fill opacity=0.92, text opacity=1, rounded corners=2pt},
  legend columns=1,
  enlarge y limits=0.08,
  nodes near coords,
  nodes near coords style={font=\tiny\bfseries, anchor=west, text=black!75},
  every axis plot/.append style={fill opacity=0.88},
  grid=both,
  grid style={gray!8, thin},
  major grid style={gray!15},
  axis line style={gray!40},
  clip=false,
]
\fill[richGreen!10] (axis cs:0,6.5) rectangle (axis cs:100,9.6);
\fill[richOrange!7] (axis cs:0,3.5) rectangle (axis cs:100,6.5);
\fill[richRed!5] (axis cs:0,-0.6) rectangle (axis cs:100,3.5);
\draw[richGreen!40, thick, dashed] (axis cs:0,6.5) -- (axis cs:100,6.5);
\draw[richRed!30, thick, dashed] (axis cs:0,3.5) -- (axis cs:100,3.5);
\node[font=\tiny\itshape, text=richGreen!60, anchor=south east, fill=white, inner sep=0.5pt] at (axis cs:98,8.8) {Structured edits};
\node[font=\tiny\itshape, text=richOrange!60, anchor=south west, fill=white, inner sep=0.5pt] at (axis cs:75,5) {Mixed};
\node[font=\tiny\itshape, text=richRed!50, anchor=south east, fill=white, inner sep=0.5pt] at (axis cs:98,1.2) {Other workflows};
\draw[richRed!35, thin, dashed] (axis cs:25,-0.6) -- (axis cs:25,9.6);
\draw[richOrange!35, thin, dashed] (axis cs:50,-0.6) -- (axis cs:50,9.6);
\draw[richGreen!35, thin, dashed] (axis cs:75,-0.6) -- (axis cs:75,9.6);
\node[font=\tiny, text=richRed!40, anchor=south] at (axis cs:25,9.7) {25\%};
\node[font=\tiny, text=richOrange!45, anchor=south] at (axis cs:50,9.7) {50\%};
\node[font=\tiny, text=richGreen!50, anchor=south] at (axis cs:75,9.7) {75\%};
\addplot[fill=richNavy!85, draw=richNavy!90!black] coordinates {(0,0) (18,1) (17,2) (38,3) (44,4) (36,5) (73,6) (75,7) (89,8) (80,9)};
\addlegendentry{Sonnet 4.5}
\addplot[fill=richOrange!80, draw=richOrange!85!black] coordinates {(10,0) (12,1) (8,2) (38,3) (22,4) (38,5) (47,6) (75,7) (89,8) (40,9)};
\addlegendentry{GPT-5.4-R}
\addplot[fill=richSteelBlue!55, draw=richSteelBlue!70!black] coordinates {(0,0) (12,1) (8,2) (25,3) (33,4) (23,5) (53,6) (63,7) (44,8) (20,9)};
\addlegendentry{GPT-5}
\end{axis}
\end{tikzpicture}
\caption{Per-category pass rates for three selected models. Shading groups structured edits, mixed categories, and other workflows. These descriptive groups do not define a read/write partition. Board results reflect an adapter without board tools.}
\label{fig:readwrite}
\end{figure}

Table~\ref{tab:per-category} and Figure~\ref{fig:readwrite} report variation across declared categories.

\begin{table}[t]
\centering
\scriptsize
\setlength{\tabcolsep}{3pt}
\begin{tabular}{@{}lrrrrr@{}}
\toprule
Category & Son. & Opus & 5.4-R & o3 & GPT-5 \\
\midrule
filters & 80 & 90 & 40 & 50 & 20 \\
folders & 89 & 56 & 89 & 44 & 44 \\
settings & 75 & 75 & 75 & 50 & 62 \\
messages-write & 73 & 67 & 47 & 47 & 53 \\
contacts & 17 & 17 & 17 & 17 & 17 \\
messages-read & 36 & 33 & 38 & 23 & 23 \\
calendar-read & 8 & 8 & 25 & 17 & 17 \\
inbox-cleanup & 44 & 22 & 22 & 22 & 33 \\
todo & 22 & 22 & 11 & 11 & 11 \\
calendar-triage & 38 & 25 & 38 & 25 & 25 \\
meeting-prep & 22 & 22 & 11 & 0 & 11 \\
multi-domain & 17 & 17 & 0 & 0 & 0 \\
calendar-write & 17 & 8 & 8 & 17 & 8 \\
cross-domain & 7 & 7 & 13 & 13 & 7 \\
orchestration & 18 & 18 & 12 & 12 & 12 \\
boards & 0 & 0 & 10 & 20 & 0 \\
\bottomrule
\end{tabular}
\caption{Per-category pass rates (percent). Some write categories have high observed rates. Board tools are absent from the evaluated adapter.}
\label{tab:per-category}
\end{table}

For Sonnet, the highest-performing categories include structured state changes: filters (80\%), folders (88.9\%), and settings (75\%).
Its calendar-write and orchestration pass rates are 16.7\% and 17.6\%, respectively. Full per-category results for all eight configurations appear in Appendix~\ref{app:full-results}.
Performance differs across assigned difficulty groups: D1 at 6--12\%, D2 at 28--42\%, and D3 at 10--24\%. These rates are not monotonic in the assigned level.
\paragraph{Recorded failure evidence.}
The available annotations do not support frequency estimates for a qualitative failure taxonomy. Appendix~\ref{app:trajectories} therefore reports only failure mechanisms grounded in recorded traces.

\paragraph{Difficulty Stratification.}
Table~\ref{tab:difficulty-results} breaks down performance by difficulty level.

\begin{table}[t]
\centering
\small
\setlength{\tabcolsep}{3pt}
\begin{tabular}{@{}lrrrrr@{}}
\toprule
Level & $n$ & Son. & 5.4-R & o3 & GPT-5 \\
\midrule
D1 & 17 & 11.8 & 5.9 & 11.8 & 11.8 \\
D2 & 121 & 42.1 & 37.2 & 30.6 & 28.1 \\
D3 & 68 & 23.5 & 17.6 & 10.3 & 10.3 \\
\bottomrule
\end{tabular}
\caption{Pass rates (percent) by assigned difficulty. Difficulty labels do not specify an exact number of calls or domains. Unavailable tools can affect low-difficulty tasks.}
\label{tab:difficulty-results}
\end{table}

\subsection{Where Models Struggle}
\label{sec:struggle}

The recorded results vary across task categories but do not support attributing failures to a single model capability.

\paragraph{Nested Object Construction.}
Calendar writes and recipient payloads have structured input requirements. Payload validity is distinct from satisfying a scenario's requested time, recipients, or other constraints. We report no nesting-depth error percentages without a documented annotation procedure.

\paragraph{Result Piping.}
Workflows require using returned identifiers in later calls. The recorded OB-N520 trace correctly creates a parent folder, uses its ID for subfolders, and then uses subfolder IDs in batch moves.

\paragraph{Temporal Reasoning.}
Time-sensitive requests depend on the fixed corpus reference time and the mock API's filtering semantics. No temporal-versus-nontemporal failure ratio is established by the available annotations.

\paragraph{Tool Call Efficiency.}
The eight recorded runs contain 378--1,782 tool calls per configuration. More calls do not guarantee a higher pass rate. Call count also reflects task mix, pagination, repeated retrieval, and zero-tool responses. Repetition alone is not evidence that a model is retrying an API error.

\paragraph{Unavailable Operations.}
There are 10 board-category scenarios and 19 scenarios declaring the boards domain. The adapter exposes no board tools. Failure on those tasks conflates tool availability with reasoning and cannot establish a board-specific capability deficit.

\begin{figure}[t]
\centering
\includegraphics[width=\columnwidth]{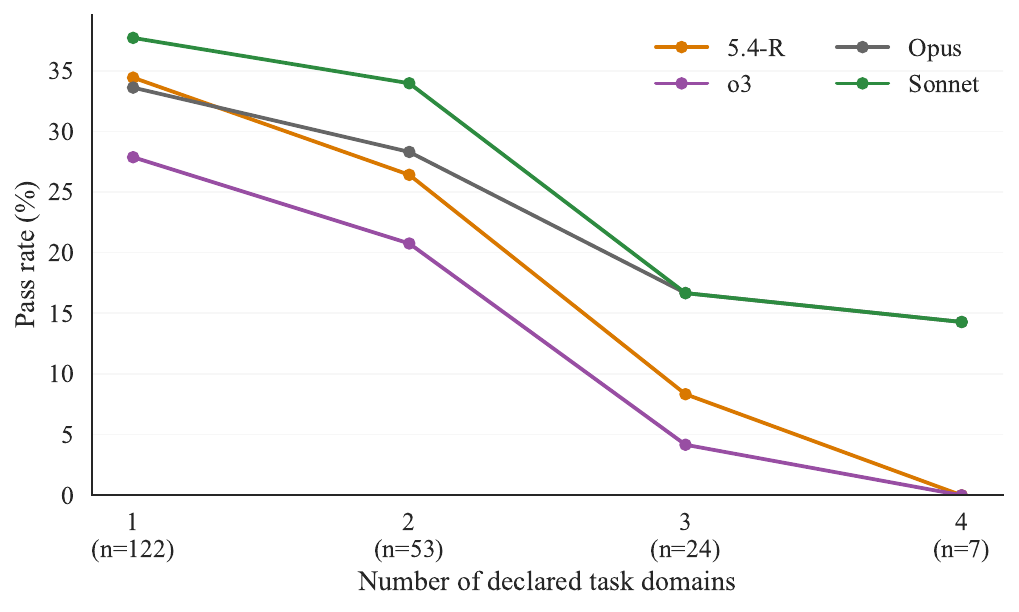}
\caption{Pass rate by declared task-domain count, joined from scenario IDs. Cohort sizes are 122, 53, 24, and 7. Differences are descriptive associations, not an experimentally isolated cost per added domain.}
\label{fig:domain-scaling}
\end{figure}

\paragraph{Domain Composition Scaling.}
Table~\ref{tab:domain-combos} and Figure~\ref{fig:domain-scaling} quantify degradation by domain count.

\begin{table}[t]
\centering
\small
\setlength{\tabcolsep}{3pt}
\begin{tabular}{@{}lrrrr@{}}
\toprule
Domains & Son. & Opus & 5.4-R & o3 \\
\midrule
1 ($n=122$) & 37.7 & 33.6 & 34.4 & 27.9 \\
2 ($n=53$) & 34.0 & 28.3 & 26.4 & 20.8 \\
3 ($n=24$) & 16.7 & 16.7 & 8.3 & 4.2 \\
4 ($n=7$) & 14.3 & 14.3 & 0.0 & 0.0 \\
\bottomrule
\end{tabular}
\caption{Pass rates (percent) by declared domain count. Cohort sizes sum to 206. Differences combine task composition, access restrictions, and model performance.}
\label{tab:domain-combos}
\end{table}

Declared domain count is associated with different task mixes and small cohorts at the high end. These comparisons do not establish a causal decline of a fixed number of percentage points per added domain.

\paragraph{Inter-Model Agreement.}
To understand whether models fail on the same scenarios or exhibit complementary strengths, we compute pairwise agreement between four explicitly named models (Table~\ref{tab:agreement}).

\begin{table}[t]
\centering
\small
\setlength{\tabcolsep}{3pt}
\begin{tabular}{@{}lrrrr@{}}
\toprule
 & Son. & Opus & 5.4-R & o3 \\
\midrule
Sonnet 4.5 & N/A & 0.711 & 0.530 & 0.420 \\
Opus 4.1 & 0.711 & N/A & 0.506 & 0.446 \\
GPT-5.4-R & 0.530 & 0.506 & N/A & 0.507 \\
o3 & 0.420 & 0.446 & 0.507 & N/A \\
\bottomrule
\end{tabular}
\caption{Jaccard similarity of recorded pass sets for the four named models. This selected cohort includes o3 and is not the top four by observed pass rate.}
\label{tab:agreement}
\end{table}

For the named Sonnet/Opus/GPT-5.4-R/o3 cohort, 24/206 scenarios pass for every model, 117/206 pass for none, and 65/206 have differing outcomes. These sets are computed by scenario ID. Appendix~\ref{app:hard-scenarios} describes the cohort failed by all four.

\paragraph{Reasoning-Labeled Configurations.}
GPT-5.4-reasoning (28.2\%) has a higher observed pass rate than GPT-5.4 (25.7\%), a difference of 2.4 percentage points before rounding.
Similarly, o3 passes 22.3\% versus GPT-5's 20.9\%, with higher mean latency (67.17 vs.\ 13.59 seconds).
GPT-5.5-reasoning uses the most tools (8.7 on average) and passes 27.7\%.
These model comparisons do not isolate planning, schema recall, tool availability, or grading effects.
Targeted interventions and a revised tool-access protocol would be needed to separate these explanations.

\paragraph{Differences Across GPT-5 Labels.}
The observed pass rates increase across GPT-5 (20.9\%), GPT-5.2 (22.8\%), and GPT-5.4 (25.7\%), a total difference of 4.9 percentage points before rounding.
All evaluated GPT-5 variants have lower observed pass rates than the two evaluated Anthropic models.
These descriptive comparisons do not establish architectural causes or statistically significant differences between models.

\subsection{Iteration Dynamics}
\label{sec:iteration}

We describe recorded iteration patterns across all eight configurations, including when passing trajectories terminate and how activity changes over a run. These retrospective summaries do not estimate performance under a different iteration budget.

\subsubsection{Retrospective Completion by Iteration}

Table~\ref{tab:iter-cap} counts recorded trajectories that both passed at the original budget and completed within $K$ iterations. This retrospective analysis is not a rerun under a lower iteration cap and does not bound such a rerun.

\begin{table}[t]
\centering
\scriptsize
\setlength{\tabcolsep}{3pt}
\begin{tabular}{@{}lrrrrrrr@{}}
\toprule
Model & 1 & 2 & 3 & 5 & 7 & 10 & 15 \\
\midrule
Sonnet 4.5 & 0.0 & 11.7 & 22.8 & 30.1 & 32.0 & 32.0 & 33.5 \\
Opus 4.1 & 0.0 & 6.8 & 15.0 & 21.8 & 25.2 & 26.7 & 29.6 \\
GPT-5.4-R & 0.0 & 7.8 & 15.0 & 22.8 & 25.2 & 27.7 & 28.2 \\
GPT-5.5-R & 0.5 & 8.7 & 13.6 & 18.4 & 21.4 & 26.2 & 27.7 \\
GPT-5.4 & 0.0 & 14.6 & 20.4 & 23.3 & 25.2 & 25.2 & 25.7 \\
GPT-5.2 & 0.5 & 11.2 & 18.0 & 20.4 & 21.8 & 22.3 & 22.8 \\
o3 & 0.5 & 4.4 & 9.7 & 14.6 & 15.0 & 19.4 & 22.3 \\
GPT-5 & 0.0 & 8.7 & 15.5 & 19.4 & 19.9 & 20.4 & 20.9 \\
\bottomrule
\end{tabular}
\caption{Retrospective completed-pass rates (percent of all 206 scenarios) at iteration $K$. No lower-budget run is simulated. Recorded final outcomes are unchanged.}
\label{tab:iter-cap}
\end{table}

Figure~\ref{fig:iter-curve} in the main paper visualizes these completion counts.

By iteration five, the share of final passes already completed ranges from 30/46 for o3 (65.2\%) to 40/43 for GPT-5 (93.0\%). Sonnet completes 62/69 passes by that point (89.9\%). Later completions contribute differently across models. These observations do not justify changing the iteration cap without a new experiment.

\subsubsection{Per-Iteration Pipeline Breakdown}
\label{sec:per-iter-pipeline}

\begin{figure*}[t]
\centering
\includegraphics[width=\textwidth]{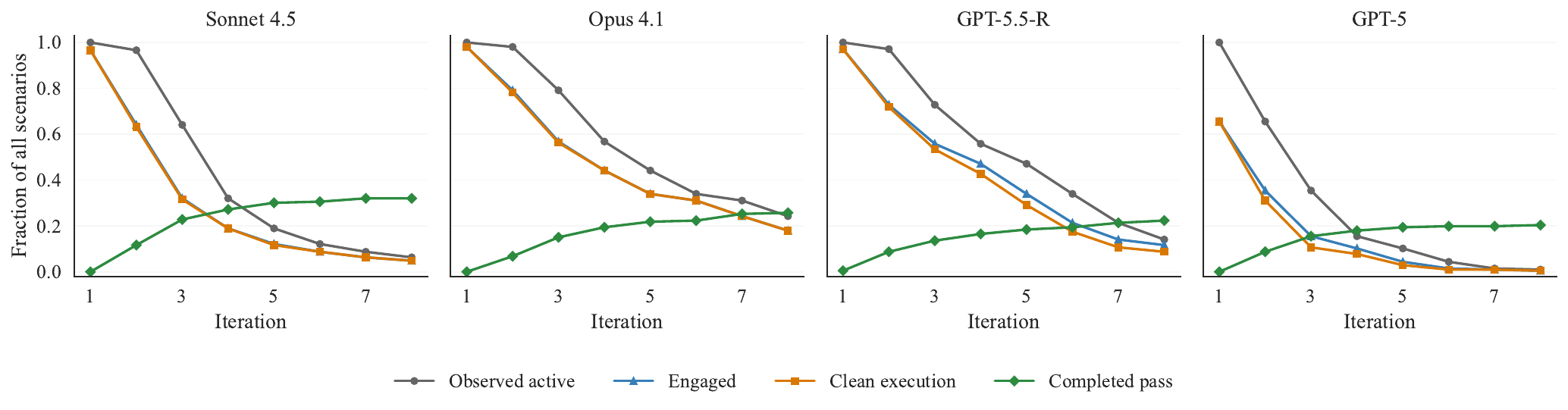}
\caption{Per-iteration activity in four recorded runs. Active denotes observed ongoing activity. Engaged denotes at least one call in that iteration. Clean execution denotes engaged trajectories with no observed API failure in that iteration. Cumulative pass counts completed passing runs. All percentages use 206 scenarios. For aborted runs with missing terminal counts, Active is an observed lower bound reconstructed from recorded calls.}
\label{fig:per-iter-funnel}
\end{figure*}

Figure~\ref{fig:per-iter-funnel} decomposes each iteration into pipeline stages, analogous to JQBench's compile$\rightarrow$execute$\rightarrow$values progression \citep{verbruggen2026jqbench}. It shows where the funnel narrows.

Activity decreases as trajectories terminate, so a smaller engaged fraction at later iterations is not itself an error. Active and engaged rates differ when a model is producing a final answer without a tool call. API failures are classified from structured response semantics, not words such as \emph{failed} appearing in retrieved messages. Four exception records have zero stored iterations despite recorded calls. Their observed activity is reconstructed from call indices, with any unrecorded terminal step left unknown.

\subsubsection{Self-Correction and Recovery}
\label{sec:recovery}

We decompose agent trajectories into four categories based on iteration count and outcome (Table~\ref{tab:recovery}).

\begin{table}[t]
\centering
\scriptsize
\setlength{\tabcolsep}{3pt}
\begin{tabular}{@{}lrrrr@{}}
\toprule
Model & 1-pass & Later pass & Fail/calls & Fail/no calls \\
\midrule
Sonnet 4.5 & 0 & 69 & 130 & 7 \\
Opus 4.1 & 0 & 61 & 141 & 4 \\
GPT-5.4-R & 0 & 58 & 138 & 10 \\
GPT-5.5-R & 1 & 56 & 144 & 5 \\
GPT-5.4 & 0 & 53 & 140 & 13 \\
GPT-5.2 & 1 & 46 & 88 & 71 \\
o3 & 1 & 45 & 136 & 24 \\
GPT-5 & 0 & 43 & 92 & 71 \\
\bottomrule
\end{tabular}
\caption{Trajectory counts partition all 206 outcomes. Iterations use observed activity when a stored count is missing. Failures are split by whether any tool call was recorded. A zero-tool response is not automatically a refusal.}
\label{tab:recovery}
\end{table}

\paragraph{Repeated tool calls.}
Repeated calls may represent pagination, retrieval of multiple entities, repeated mutations, or retries. Table~\ref{tab:retry} counts failed trajectories with calls that contain adjacent calls to the same tool with different serialized arguments. This observable pattern alone does not establish that an error occurred or that a model recognized it.

\begin{table}[t]
\centering
\scriptsize
\setlength{\tabcolsep}{3pt}
\begin{tabular}{@{}lrrr@{}}
\toprule
Model & Fail/calls & Repeated & \% \\
\midrule
Sonnet 4.5 & 130 & 72 & 55.4 \\
Opus 4.1 & 141 & 93 & 66.0 \\
GPT-5.4-R & 138 & 71 & 51.4 \\
GPT-5.5-R & 144 & 101 & 70.1 \\
GPT-5.4 & 140 & 40 & 28.6 \\
GPT-5.2 & 88 & 21 & 23.9 \\
o3 & 136 & 105 & 77.2 \\
GPT-5 & 92 & 26 & 28.3 \\
\bottomrule
\end{tabular}
\caption{Failed trajectories with adjacent same-tool calls using different arguments. The denominator is failed trajectories with recorded calls. This is a repetition measure, not an error-recovery rate.}
\label{tab:retry}
\end{table}

\paragraph{Zero-tool responses.}
Table~\ref{tab:recovery} separates failed responses with no recorded calls. Their content may be a refusal, an unsupported answer, or a report of missing capabilities. Those interpretations require inspecting the response rather than inferring intent from call count.

\subsubsection{Pass vs.\ Fail Behavioral Signatures}
\label{sec:pass-fail}

Table~\ref{tab:pass-fail-tc} reveals distinct behavioral patterns between passing and failing trajectories.

\begin{table}[t]
\centering
\scriptsize
\setlength{\tabcolsep}{3pt}
\begin{tabular}{@{}lrrrr@{}}
\toprule
Model & Calls/pass & Calls/fail & Iter/pass & Iter/fail \\
\midrule
Sonnet 4.5 & 5.28 & 5.88 & 3.61 & 3.69 \\
Opus 4.1 & 4.59 & 5.70 & 4.75 & 5.85 \\
GPT-5.4-R & 3.00 & 3.84 & 4.00 & 4.80 \\
GPT-5.5-R & 7.60 & 9.05 & 4.81 & 4.56 \\
GPT-5.4 & 2.70 & 2.64 & 2.98 & 3.12 \\
GPT-5.2 & 3.34 & 1.42 & 3.13 & 2.10 \\
o3 & 4.63 & 7.06 & 5.63 & 7.79 \\
GPT-5 & 3.02 & 1.52 & 3.21 & 2.12 \\
\bottomrule
\end{tabular}
\caption{Mean recorded tool calls and observed iterations, grouped by outcome. The groups differ in task composition. These differences are descriptive, not evidence of a causal effect of extra calls.}
\label{tab:pass-fail-tc}
\end{table}

Passing and failing cohorts contain different scenarios. Additional calls can reflect useful work, repetition, or harder tasks, while zero-tool failures lower the failure-group mean. The table does not identify an optimal tool-use policy.

\subsubsection{Iteration Efficiency by Bucket}
\label{sec:iter-bucket}

We group scenarios by the number of iterations the agent used and compute per-bucket pass rates (Table~\ref{tab:iter-bucket}).

\begin{table}[t]
\centering
\scriptsize
\setlength{\tabcolsep}{3pt}
\begin{tabular}{@{}lccccc@{}}
\toprule
Model & 1 & 2 & 3--4 & 5--7 & 8+ \\
\midrule
Sonnet 4.5 & 0/7 & 24/67 & 32/93 & 10/26 & 3/13 \\
Opus 4.1 & 0/4 & 14/39 & 26/72 & 12/41 & 9/50 \\
GPT-5.4-R & 0/11 & 16/47 & 27/74 & 9/41 & 6/33 \\
GPT-5.5-R & 1/6 & 17/50 & 16/53 & 10/68 & 13/29 \\
GPT-5.4 & 0/13 & 30/91 & 17/71 & 5/25 & 1/6 \\
GPT-5.2 & 1/72 & 22/69 & 17/50 & 5/10 & 2/5 \\
o3 & 1/25 & 8/18 & 16/37 & 6/38 & 15/88 \\
GPT-5 & 0/71 & 18/62 & 19/52 & 4/19 & 2/2 \\
\bottomrule
\end{tabular}
\caption{Recorded passes and scenario counts by observed iteration bucket. Small high-iteration cohorts can have unstable rates. These are not rerun results at different budgets.}
\label{tab:iter-bucket}
\end{table}

Bucket membership is determined after execution. It cannot establish an optimal iteration budget, because models and scenarios enter the buckets for different reasons.

\subsubsection{Score Distribution}
\label{sec:score-dist}

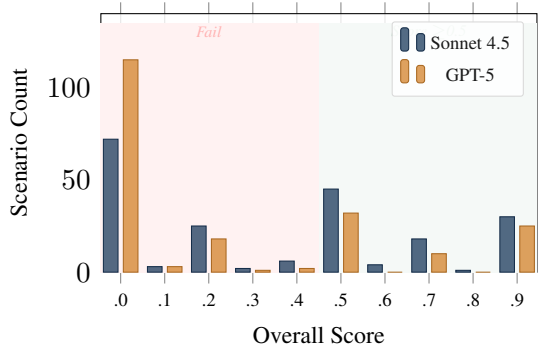
\begin{figure}[t]
\centering
\begin{tikzpicture}
\begin{axis}[
  ybar,
  width=\columnwidth-10pt,
  height=5cm,
  bar width=5.5pt,
  ymin=0, ymax=140,
  ylabel={Scenario Count},
  xlabel={Overall Score},
  ylabel style={font=\small},
  xlabel style={font=\small},
  xtick={0.05,0.15,0.25,0.35,0.45,0.55,0.65,0.75,0.85,0.95},
  xticklabels={.0,.1,.2,.3,.4,.5,.6,.7,.8,.9},
  xticklabel style={font=\scriptsize},
  legend style={at={(0.97,0.97)}, anchor=north east, font=\scriptsize, draw=black!15, fill=white, fill opacity=0.9, rounded corners=1pt},
  enlarge x limits=0.05,
  every axis plot/.append style={fill opacity=0.75},
  grid=both,
  grid style={gray!12, thin},
  major grid style={gray!20},
]
\draw[red!60, thick, dashed] (axis cs:0.5,0) -- (axis cs:0.5,135);
\node[font=\tiny, text=red!60, anchor=south, rotate=90] at (axis cs:0.49,60) {Score threshold};
\fill[red!5] (axis cs:0,0) rectangle (axis cs:0.5,135);
\fill[richGreen!5] (axis cs:0.5,0) rectangle (axis cs:1.0,135);
\node[font=\tiny, text=red!30, anchor=center] at (axis cs:0.25,130) {\textit{Fail}};
\node[font=\tiny, text=richGreen!50, anchor=center] at (axis cs:0.75,130) {\textit{Score $\geq$0.5}};
\addplot[fill=richNavy, draw=richNavy!80!black] coordinates {(0.05,72) (0.15,3) (0.25,25) (0.35,2) (0.45,6) (0.55,45) (0.65,4) (0.75,18) (0.85,1) (0.95,30)};
\addlegendentry{Sonnet 4.5}
\addplot[fill=richOrange, draw=richOrange!80!black] coordinates {(0.05,115) (0.15,3) (0.25,18) (0.35,1) (0.45,2) (0.55,32) (0.65,0) (0.75,10) (0.85,0) (0.95,25)};
\addlegendentry{GPT-5}
\end{axis}
\end{tikzpicture}
\caption{Score distribution for Sonnet and GPT-5 across 206 scenarios. Bins have width 0.1, and the last includes both endpoints [0.9, 1.0]. Shading separates scores below 0.5 from those at or above it. The latter can still fail critical assertions. Both distributions are bimodal, with many scenarios near 0 or above 0.5 and few intermediate cases.}
\label{fig:score-dist}
\end{figure}

Figure~\ref{fig:score-dist} shows recorded overall scores. A score of at least 0.5 is necessary but not sufficient to pass because critical assertions are also required.

Scores alone do not distinguish API failure, incomplete work, or critical-check failure. The recorded OB-H307 example reaches 0.50 yet fails its critical LLM rubric.

\subsection{Graduated Execution Pipeline}
\label{sec:graduated}

Table~\ref{tab:graduated} reports separate scenario-level indicators. These are not all nested stages: a trajectory can produce an answer after an API error, and a stored pass can occur without a recorded call. Clean execution is an API-behavior indicator, not evidence of task correctness.

\begin{table}[t]
\centering
\scriptsize
\input{figures/out/table_execution_pipeline}
\caption{Scenario indicators from recorded runs (percent of 206). Engagement requires a recorded tool call. Clean execution additionally requires no observed API failure. Answer excludes runner error and max-iteration sentinel strings. Pass uses the recorded grade.}
\label{tab:graduated}
\end{table}

Execution classification uses top-level error envelopes, nonempty batch-failure lists, and failed mutation return values. This avoids treating text such as \texttt{failed: []}, nullable calendar fields, or message bodies containing error-related words as failures. The classification affects execution statistics but does not recompute task grades.

\subsection{Tool Call Success and Error Feedback}
\label{sec:tool-success}

\subsubsection{Execution Classification}
Nonempty top-level error envelopes and batch failure lists count as observed API failures. A mutation returning \texttt{false} or \texttt{null} also counts as a failure. An absent-object lookup is recorded separately from mutation failure. These rules measure the API result, not whether arguments matched the user's intended task.

Tool results were truncated at 8,000 characters. A truncated array is not classified as a failure merely because it cannot be parsed as complete JSON. The analysis records truncation and distinguishes unclassifiable results from observed success or failure. Successful calls can still return incomplete data, use the wrong filter, or perform an unintended operation.

\begin{table}[t]
\centering
\scriptsize
\input{figures/out/table_tool_success}
\caption{Tool-call execution statistics using structured failure evidence. Calls are individual invocations. First-call rates use trajectories with at least one call. Stored grading outcomes are not recomputed.}
\label{tab:tool-success}
\end{table}

For example, Sonnet has four observed failures among 1,170 calls, a 99.66\% rate of calls without observed API failure. Its much lower scenario pass rate demonstrates the distinction between invoking an API and completing a task. The former claim that every Sonnet batch move failed arose from treating an empty \texttt{failed} array as an error.

\subsubsection{Error Feedback Utilization}
For an observed failure followed by another call in the same trajectory, Table~\ref{tab:feedback} records whether that next call executes successfully and whether it uses the same tool. Calls issued together in one model iteration cannot establish response to feedback. These descriptive next-call rates must not be interpreted as a causal measure of learning from an error.

\begin{table}[t]
\centering
\scriptsize
\input{figures/out/table_recovery}
\caption{Next-call behavior after observed API failures. Denominators exclude terminal failures with no following call. Same-tool continuation and successful next execution do not by themselves establish recovery of the original task.}
\label{tab:feedback}
\end{table}

With few observed failures for some models, recovery fractions have small denominators. The data do not establish that late execution errors or repeated attempts explain most task failures.

\subsection{Complexity Correlations}
\label{sec:complexity}

\begin{figure*}[t]
\centering
\includegraphics[width=\textwidth]{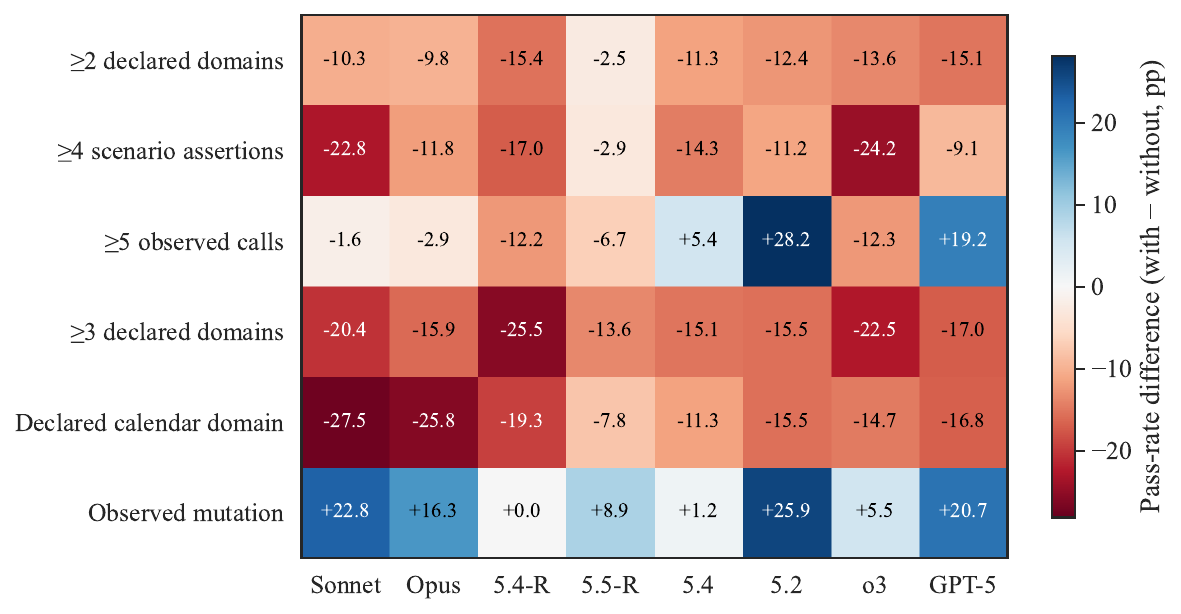}
\caption{Signed pass-rate differences (percentage points) for documented complexity proxies. Scenario metadata are joined by ID, and execution-derived features describe observed behavior. Cohort counts and feature definitions accompany the regenerated analysis.}
\label{fig:complexity-heatmap}
\end{figure*}

Figure~\ref{fig:complexity-heatmap} joins every recorded result to its scenario definition by ID before forming feature cohorts. Each signed difference is the pass rate for scenarios with the feature minus the rate without it. Signs are preserved. These are descriptive associations among overlapping cohorts, not controlled causal effects.

Declared domain count, assigned difficulty, assertion count, and observed tool-call count describe different aspects of a trajectory. In particular, observed call count is not a known minimum number of calls required. Task categories and available tools can confound these comparisons.

\begin{figure}[t]
\centering
\includegraphics[width=\columnwidth]{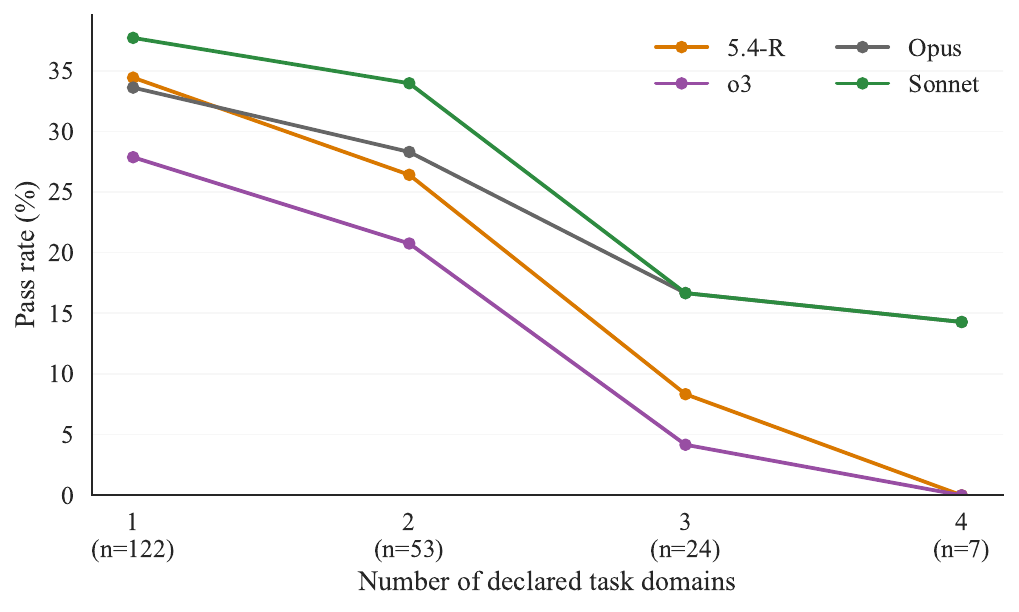}
\caption{Pass rates grouped by the number of declared task domains: 122, 53, 24, and 7 scenarios. The grouping does not imply a prescribed order of API calls.}
\label{fig:pass-by-domains}
\end{figure}

Figure~\ref{fig:pass-by-domains} presents the declared-domain cohorts directly. The four-domain group contains only seven scenarios. No fixed per-domain penalty or causal claim about assertion count is inferred.

\section{Supplementary Analysis}
\label{sec:analysis}

\subsection{Structural Requirements in Email Tasks}

The benchmark represents three recurring structural requirements:

\paragraph{Nested entity structures.}
Unlike flat key-value stores, email entities have deeply nested types.
A \texttt{Message} has \texttt{from.emailAddress.address} (3 levels), \texttt{flag.flagStatus}, and \texttt{Recipient[]} arrays.
Calendar events add \texttt{start.dateTime}/\texttt{start.timeZone} and \texttt{EventAttendee[]} with nested \texttt{status.response} fields.
An incorrect field path can make a call invalid or direct it to the wrong value.

\paragraph{Temporal reasoning.}
Many scenarios require date arithmetic relative to the reference time and understanding temporal relationships (``tomorrow'', ``last 3 days'', ``within 30 minutes'').

\paragraph{Multi-step dependencies.}
Write workflows have ordering constraints, such as creating a folder before moving emails or retrieving an attendee list before forwarding.
Failure at one step can propagate through the remaining workflow.

\subsection{Implications for Agent Design}

The recorded failures motivate several interventions for future evaluation:

\textbf{Schema grounding.}
Explicit schema documentation and structured output constraints could help with payload construction, but their effect needs separate evaluation.

\textbf{Pipeline planning.}
Explicit dependency planning is a candidate intervention for incomplete workflows. Its effect is not measured here.
A ``plan-then-execute'' architecture that first generates a dependency DAG (e.g., ``folder ID from step~1 feeds step~3'') could address this.

\textbf{Typed validation.}
Runtime type checking or schema validation before API dispatch could identify malformed payloads. Whether such a layer improves task completion remains an empirical question.

\textbf{Domain-specialized prompting.}
Prompting interventions require evaluation with an explicit tool manifest. Board tools must first be made available before testing whether prompting improves board-task performance.

\subsection{Recorded Resource Use}
The result artifacts do not include token-usage data. Table~\ref{tab:tokens} therefore reports recorded calls, observed iterations, and wall-clock latency.

\begin{table}[t]
\centering
\small
\begin{tabular}{@{}lrrr@{}}
\toprule
Model & Calls & Iterations & Latency (s) \\
\midrule
Sonnet 4.5 & 5.68 & 3.66 & 27.06 \\
Opus 4.1 & 5.37 & 5.52 & 40.56 \\
GPT-5.4-R & 3.61 & 4.58 & 21.84 \\
GPT-5.5-R & 8.65 & 4.63 & 31.69 \\
GPT-5.4 & 2.66 & 3.09 & 17.80 \\
GPT-5.2 & 1.86 & 2.33 & 15.84 \\
o3 & 6.51 & 7.31 & 67.17 \\
GPT-5 & 1.83 & 2.35 & 13.59 \\
\bottomrule
\end{tabular}
\caption{Per-scenario means from recorded runs. Observed iterations use recorded call indices where a stored iteration count is missing. Exact unrecorded terminal activity remains unknown.}
\label{tab:tokens}
\end{table}

\subsection{Efficiency--Accuracy Tradeoff}
\label{sec:efficiency-detail}

Figure~\ref{fig:efficiency} in the main paper plots the empirical latency--accuracy frontier. No other configuration is both at least as accurate and at least as fast as a frontier point, with one strict improvement.

Efficiency is defined here as pass-rate percentage points divided by mean scenario latency in seconds. Using unrounded recorded latencies, GPT-5 has a ratio of 1.5355 pp/s, GPT-5.4-R has 1.2890 pp/s, and Sonnet has 1.2377 pp/s. This ratio is one preference rule, not a universal ranking.

Sonnet has the highest observed accuracy and is on the Pareto frontier. There is no unique best frontier model without specifying an accuracy/latency tradeoff. o3's mean latency is 67.1671 s versus Sonnet's 27.0616 s, a 2.48$\times$ ratio.

\subsection{Corpus Realism vs.\ Scale}

Our seed corpus (65 messages) is deliberately small compared to production mailboxes.
\textbf{Advantages}: fixed mock data, local API operations, and high interconnection density. Model inference and LLM grading are not deterministic.
\textbf{Limitations}: limited pagination pressure, reduced needle-in-haystack difficulty, and no rate limiting.
Future work could introduce scaled variants with 1,000+ messages.

\subsection{Variance and Reproducibility}
The analysis contains one complete result file per reported configuration. The available artifacts do not provide multiple runs per configuration, so we do not report replicate variance or attribute observed variance to the judge.

Static checking is deterministic for fixed inputs, but model outputs and LLM judgments may vary. Figure~\ref{fig:main-results}'s Wilson intervals quantify the separate, explicitly stated scenario-sampling uncertainty.

\section{Protocol-Specific Constraints}
\label{sec:limitations}

\textbf{Simulated environment.}
The mock API cannot capture network latency, rate limiting, eventual consistency, or OAuth authentication flows present in production deployments.

\textbf{Corpus scale.}
The seed corpus ($\sim$65 messages) is smaller than production mailboxes (which may have thousands of messages), limiting evaluation of pagination handling, search precision at scale, and information retrieval over large collections.

\textbf{Single-turn evaluation.}
Real users iteratively refine agent outputs through multi-turn correction (``no, I meant next Tuesday'').  Our single-turn protocol does not capture this interactive loop.

\textbf{English-only.}
All corpus content and queries are in English.  Multilingual email management (common in global enterprises) is not evaluated.

\textbf{Single-user baseline.}
All reported scenarios run against Kaminski's corpus, including tasks naming another user. Cross-persona generalization is not evaluated. Missing board, directory, and contact-group tools limit capability attribution for failures on those operations.

\textbf{Grader limitations.}
LLM judgments may vary and miss domain-specific nuances. Available artifacts do not establish repeated-run stability or expert agreement. Equal weighting and critical checks do not remove this uncertainty. Answer-pattern and snapshot-check limitations also affect validity.

\textbf{No multi-agent evaluation.}
We evaluate single-agent performance only.  Multi-agent delegation patterns (e.g., a planning agent dispatching to specialist sub-agents) may achieve different results.

\section{Data Provenance Details}
\label{sec:ethics}

EmailBench uses fabricated content inspired by publicly available Enron email data \citep{klimt2004enron}.
It includes no original email bodies or private mailbox exports. Message bodies, calendar events, tasks, and other scenario content are synthetic. Publicly documented names, email addresses, roles, and organizational relationships remain, so the corpus should not be described as containing no real personal information.
The Enron corpus was released by FERC under public disclosure mandate and has been widely used in email research \citep{klimt2004enron, zhang2019email}. Public availability alone does not establish permission to reuse or redistribute every source item or identifier.

The available benchmark artifacts expose only aggregate intent-frequency counts from the prototype telemetry referenced in \S\ref{sec:construction}. They do not establish how the source telemetry was collected, de-identified, or retained, so we make no stronger privacy claim about the unavailable source records.

\section{Reproducibility and Availability}
\label{sec:reproducibility}

The benchmark environment is self-contained, while model and judge inference requires external endpoints. The available artifact supports the following forms of inspection and replay:

\begin{itemize}[leftmargin=*,itemsep=2pt]
\item \textbf{Self-contained state.} The mock API runs in memory over a frozen corpus and does not require a live email provider.
\item \textbf{Deterministic corpus.} The seed corpus is a typed constant, and corpus initialization is deterministic.
\item \textbf{Grading scripts.} The 258 static assertions are encoded as TypeScript predicates or answer checks. The 211 LLM rubrics and the judge prompt are versioned alongside the benchmark.
\item \textbf{Agent adapter.} The adapter exposes the evaluated tool schemas and permits replacement of the model endpoint without changing benchmark state.
\item \textbf{Recorded evidence.} Result files preserve final answers, abbreviated tool traces, grades, internal endpoint IDs, run IDs, and timestamps. They do not preserve an exact source-commit and tool-schema manifest. Offline validation reproduces the stored component scores, combined scores, and critical-check pass flags for all 1,648 saved scenario results.
\end{itemize}

The reviewed repository contains the benchmark code, corpus, scenarios, grading scripts, and baseline agent implementations. An explicit license covering every artifact was not located, so we make no blanket licensing claim. Model and judge execution also requires access to the configured hosted services and credentials.

\section{Future Evaluation}
\label{sec:future-evaluation}

Future evaluations should separate environment access from model capability by exposing every operation required by the scenarios and selecting the declared user corpus. Repeated runs and expert validation of the LLM judge are also needed. Additional benchmark variants could study larger corpora, multi-turn interaction, provider-specific implementations, schema-grounding interventions, and multi-agent execution. These extensions require new experiments and are not implied by the reported baseline.


\section{Batch~4 Gap-Filling Details}
\label{app:gap-fill}

After the first three batches (122 scenarios), we compute per-category coverage and generate 84 additional scenarios.  Table~\ref{tab:gap-fill-app} shows the targeted additions.

\begin{table}[H]
\centering
\small
\begin{tabular}{@{}lrrl@{}}
\toprule
\textbf{Category} & \textbf{Before} & \textbf{Added} & \textbf{Final} \\
\midrule
meeting-prep       & 1  & 8 & 9  \\
calendar-triage    & 2  & 6 & 8  \\
settings           & 2  & 6 & 8  \\
folders            & 3  & 6 & 9  \\
inbox-cleanup      & 3  & 6 & 9  \\
calendar-write     & 7  & 5 & 12 \\
contacts           & 7  & 5 & 12 \\
todo               & 4  & 5 & 9  \\
boards             & 5  & 5 & 10 \\
filters            & 5  & 5 & 10 \\
\emph{+ 6 other categories} & N/A & 27 & N/A \\
\bottomrule
\end{tabular}
\caption{Gap-filling in Batch~4. Categories with fewer than eight scenarios received targeted additions to establish minimum category coverage.}
\label{tab:gap-fill-app}
\end{table}

\section{Static Assertion Distribution}
\label{app:assertions}

Across 206 scenarios, there are 258 static assertions (mean 1.252). Forty-three scenarios have none. Counts are regenerated from scenario definitions and checked against the recorded assertion outcomes.

\begin{table}[t]
\centering
\small
\setlength{\tabcolsep}{3pt}
\begin{tabular}{@{}lrr@{}}
\toprule
Type & Count & \% \\
\midrule
\texttt{return-contains} & 123 & 47.7 \\
\texttt{state-check} & 101 & 39.1 \\
\texttt{exists} & 32 & 12.4 \\
\texttt{sent-message} & 2 & 0.8 \\
\texttt{count} & 0 & 0.0 \\
\texttt{not-exists} & 0 & 0.0 \\
\texttt{property} & 0 & 0.0 \\
\texttt{moved-message} & 0 & 0.0 \\
\texttt{deleted} & 0 & 0.0 \\
\textbf{Total} & \textbf{258} & 100.0 \\
\bottomrule
\end{tabular}
\caption{Code-derived distribution of 258 static assertions. Five supported types have no instances, and four types are used.}
\label{tab:assertion-dist-app}
\end{table}

\clearpage
\twocolumn[{%
\section{Full Per-Category Results}
\label{app:full-results}

Table~\ref{tab:full-category} presents pass rates across all 16 categories for all eight configurations.

\begin{center}
\small
\setlength{\tabcolsep}{3pt}
\begin{tabular}{@{}lrrrrrrrr@{}}
\toprule
Category & Son. & Opus & 5.4-R & 5.5-R & 5.4 & 5.2 & o3 & GPT-5 \\
\midrule
filters & 80.0 & 90.0 & 40.0 & 40.0 & 30.0 & 50.0 & 50.0 & 20.0 \\
folders & 88.9 & 55.6 & 88.9 & 66.7 & 55.6 & 44.4 & 44.4 & 44.4 \\
settings & 75.0 & 75.0 & 75.0 & 50.0 & 62.5 & 62.5 & 50.0 & 62.5 \\
messages-write & 73.3 & 66.7 & 46.7 & 60.0 & 46.7 & 53.3 & 46.7 & 53.3 \\
contacts & 16.7 & 16.7 & 16.7 & 16.7 & 25.0 & 8.3 & 16.7 & 16.7 \\
messages-read & 35.9 & 33.3 & 38.5 & 25.6 & 35.9 & 28.2 & 23.1 & 23.1 \\
calendar-read & 8.3 & 8.3 & 25.0 & 25.0 & 16.7 & 8.3 & 16.7 & 16.7 \\
inbox-cleanup & 44.4 & 22.2 & 22.2 & 55.6 & 11.1 & 22.2 & 22.2 & 33.3 \\
todo & 22.2 & 22.2 & 11.1 & 0.0 & 0.0 & 0.0 & 11.1 & 11.1 \\
calendar-triage & 37.5 & 25.0 & 37.5 & 37.5 & 37.5 & 25.0 & 25.0 & 25.0 \\
meeting-prep & 22.2 & 22.2 & 11.1 & 22.2 & 22.2 & 11.1 & 0.0 & 11.1 \\
multi-domain & 16.7 & 16.7 & 0.0 & 0.0 & 0.0 & 8.3 & 0.0 & 0.0 \\
calendar-write & 16.7 & 8.3 & 8.3 & 25.0 & 16.7 & 16.7 & 16.7 & 8.3 \\
cross-domain & 6.7 & 6.7 & 13.3 & 26.7 & 20.0 & 13.3 & 13.3 & 6.7 \\
orchestration & 17.6 & 17.6 & 11.8 & 11.8 & 17.6 & 11.8 & 11.8 & 11.8 \\
boards & 0.0 & 0.0 & 10.0 & 0.0 & 0.0 & 0.0 & 20.0 & 0.0 \\
\bottomrule
\end{tabular}
\captionof{table}{Pass rates by declared scenario category for all eight configurations. Rows are regenerated by joining recorded outcomes to scenario IDs.}
\label{tab:full-category}
\end{center}
\vspace{1em}
}]

\section{Complete Scenario Examples}
\label{app:scenario-examples}
These queries and rubrics come from the evaluated scenario definitions. Static checks below describe executable conditions rather than relying only on their labels.

\subsection{OB-H312: Sender Percentages}
\textbf{Category:} messages-read. \textbf{Difficulty:} D2. \textbf{Domains:} messages.

\textbf{Query:} \emph{For each person who emailed me, calculate what percentage of their emails to me are unread. Show only people with $>$50\% unread rate, sorted by unread percentage descending.}

\textbf{Static assertions.} Both static assertions are critical. The answer must contain the literal string \lstinline!\d+(\.\d+)?%! and the case-insensitive substring \emph{lay}. Although the first pattern has regex syntax, the executed checker uses substring matching. This is a grading limitation, not a desired output format. The recorded Sonnet run makes two message-list calls in three iterations and scores 0. The percentage assertion fails despite an answer containing 100.0\%. Its LLM judgment separately finds the sender analysis incorrect.

\textbf{LLM rubric (critical, correctness).} Must compute unread/total ratio per sender from inbox messages. Senders with $>$50\% unread: Lay (3 emails, most unread since appearing at hoursAgo 9,46,70), Belden (2 emails both likely unread), Mark (2 emails). Must show exact percentages and sort descending.

\subsection{OB-N520: Folder Hierarchy}
\textbf{Category:} folders. \textbf{Difficulty:} D3. \textbf{Domains:} folders, messages.

\textbf{Query:} \emph{Create a folder hierarchy: 'Projects' with subfolders 'Raptor', 'Dynegy', and 'FERC'. Then move all emails mentioning each project keyword to the appropriate subfolder.}

\textbf{Static assertions.} All five static assertions are critical: folders named Projects, Raptor, Dynegy, and FERC must exist, and at least three move records must exist. These predicates check names and move count, not parent-child relationships or keyword routing.

\textbf{LLM rubric (critical, action-taken).} Must create parent 'Projects' folder, then 3 subfolders under it. Then scan all emails for 'Raptor', 'Dynegy', 'FERC' keywords in subject/body and move matches to the right subfolder. Report move counts per folder.

\subsection{OB-H307: Ordered Inbox Triage}
\textbf{Category:} inbox-cleanup. \textbf{Difficulty:} D3. \textbf{Domains:} messages, folders.

\textbf{Query:} \emph{Set up an email triage system: (1) Create folders 'Action Required', 'FYI Only', and 'Automated'. (2) Move all emails from automated senders (enrononline, legal-tickets, gasbank) to 'Automated'. (3) Move all high-importance emails to 'Action Required'. (4) Move all remaining unread emails to 'FYI Only'. Do NOT move any emails that are already in Sent Items.}

\textbf{Static assertions.} All six static assertions are critical. Folders named Action Required, FYI Only, and Automated must exist. At least four move records must involve the selected automated IDs (msg-003/016/024/039/007/012), and at least four must involve selected high-importance IDs (msg-001/006/008/013/027/036). None may involve Sent Items IDs msg-041 through msg-055. The two move-count checks do not test destinations. Ordering, remaining unread messages, and reporting are judged by the rubric.

\textbf{LLM rubric (critical, action-taken).} Steps must execute in order: automated first (so msg-024/msg-039 go to Automated despite being high-importance), then high-importance, then remaining unread. Sent items must be excluded. Must report counts per folder.

\section{Agent System Prompt}
\label{app:system-prompt}

All eight reported configurations receive the identical system prompt below, with no few-shot examples, chain-of-thought instructions, or tool documentation beyond the JSON schema definitions:

\begin{lstlisting}[language={},basicstyle=\ttfamily\scriptsize,frame=single,breaklines=true,columns=flexible]
You are an AI assistant that manages Microsoft
Outlook. You have access to tools for managing
emails, calendar, contacts, folders, rules,
todo lists, and settings.

When the user asks you to do something:
1. Think about which tools you need to call
2. Call the appropriate tools to gather data
   or perform actions
3. Provide a clear, complete answer based on
   the tool results

Guidelines:
- For read queries: call the relevant
  list/search/get tools, then summarize the
  results clearly
- For write actions: call the appropriate
  create/update/delete/send tools
- For multi-step tasks: break them down and
  execute each step
- Always provide specific data (counts, names,
  dates) from tool results -- never make up
  information
- When listing items, include key details
  (subjects, senders, dates, etc.)
\end{lstlisting}

The prompt is deliberately minimal (148 tokens) to characterize behavior without task-specific prompt engineering. The agent receives no information about the corpus, entity IDs, API schemas beyond tool-definition JSON, or expected output format.

\paragraph{Agent Loop Configuration.}
Maximum iterations: 15. The checked-in agent request omits temperature and sets \texttt{max\_completion\_tokens=4096}. Its 180-second bridge subprocess limit applies per request, not per scenario. Tool results are truncated to 8,000 characters. When no tool calls are returned by the model, the loop terminates and the text response is taken as the final answer. Judge requests use \texttt{max\_completion\_tokens=2048} and a 120-second subprocess limit. The underlying bridge uses a 120-second HTTP timeout.

\clearpage
\section{Complete Evaluated Tool Surface}
\label{app:api-surface}
The adapter exposes 46 function tools with JSON-schema parameters. Table~\ref{tab:full-api} is regenerated from these definitions. The 119-method operational API contains additional functionality, but no board, directory, or contact-group tool is supplied in the reported runs.

\begin{table*}[t]
\centering
\scriptsize
\setlength{\tabcolsep}{3pt}
\begin{tabularx}{\textwidth}{@{}p{4.8cm}p{1.25cm}c>{\raggedright\arraybackslash}X@{}}
\toprule
Tool name & Domain & R/W & Parameters \\
\midrule
\texttt{messages\_\allowbreak list} & messages & R & top, filter, orderBy, skip, folderId \\
\texttt{messages\_\allowbreak get} & messages & R & id \\
\texttt{messages\_\allowbreak search} & messages & R & query, top \\
\texttt{messages\_\allowbreak count} & messages & R & filter \\
\texttt{messages\_\allowbreak send} & messages & W & subject, body, toRecipients, ccRecipients, importance \\
\texttt{messages\_\allowbreak createDraft} & messages & W & subject, body, toRecipients \\
\texttt{messages\_\allowbreak update} & messages & W & id, patch \\
\texttt{messages\_\allowbreak move} & messages & W & id, destinationFolderId \\
\texttt{messages\_\allowbreak delete} & messages & W & id \\
\texttt{messages\_\allowbreak reply} & messages & W & id, comment \\
\texttt{messages\_\allowbreak replyAll} & messages & W & id, comment \\
\texttt{messages\_\allowbreak forward} & messages & W & id, toRecipients, comment \\
\texttt{messages\_\allowbreak batchMove} & messages & W & ids, destinationFolderId \\
\texttt{folders\_\allowbreak list} & folders & R & N/A \\
\texttt{folders\_\allowbreak get} & folders & R & id \\
\texttt{folders\_\allowbreak getByName} & folders & R & name \\
\texttt{folders\_\allowbreak create} & folders & W & displayName, parentFolderId \\
\texttt{folders\_\allowbreak rename} & folders & W & id, newName \\
\texttt{folders\_\allowbreak delete} & folders & W & id \\
\texttt{filters\_\allowbreak list} & filters & R & N/A \\
\texttt{filters\_\allowbreak create} & filters & W & name, isEnabled, criteria, actions \\
\texttt{filters\_\allowbreak update} & filters & W & id, patch \\
\texttt{filters\_\allowbreak delete} & filters & W & id \\
\texttt{calendar\_\allowbreak list} & calendar & R & top, startDateTime, endDateTime \\
\texttt{calendar\_\allowbreak get} & calendar & R & id \\
\texttt{calendar\_\allowbreak create} & calendar & W & subject, start, end, location, attendees, body, isAllDay \\
\texttt{calendar\_\allowbreak update} & calendar & W & id, patch \\
\texttt{calendar\_\allowbreak delete} & calendar & W & id \\
\texttt{calendar\_\allowbreak accept} & calendar & W & id, comment \\
\texttt{calendar\_\allowbreak decline} & calendar & W & id, comment \\
\texttt{contacts\_\allowbreak list} & contacts & R & top \\
\texttt{contacts\_\allowbreak search} & contacts & R & query \\
\texttt{contacts\_\allowbreak create} & contacts & W & displayName, givenName, surname, emailAddresses, companyName, department, jobTitle \\
\texttt{contacts\_\allowbreak delete} & contacts & W & id \\
\texttt{todo\_\allowbreak lists} & todo & R & N/A \\
\texttt{todo\_\allowbreak createList} & todo & W & displayName \\
\texttt{todo\_\allowbreak tasks} & todo & R & listId \\
\texttt{todo\_\allowbreak createTask} & todo & W & listId, title, body, dueDateTime, importance \\
\texttt{todo\_\allowbreak updateTask} & todo & W & listId, taskId, patch \\
\texttt{todo\_\allowbreak deleteTask} & todo & W & listId, taskId \\
\texttt{settings\_\allowbreak getVacationResponder} & settings & R & N/A \\
\texttt{settings\_\allowbreak setVacationResponder} & settings & W & enabled, message, startDateTime, endDateTime, sendToContactsOnly \\
\makecell[l]{\texttt{settings\_}\\\texttt{getSenderClassifications}} & settings & R & N/A \\
\makecell[l]{\texttt{settings\_}\\\texttt{createSenderClassification}} & settings & W & senderEmailAddress, classifyAs \\
\texttt{settings\_\allowbreak getLabels} & settings & R & N/A \\
\texttt{getCurrentUser} & identity & R & N/A \\
\bottomrule
\end{tabularx}
\caption{All 46 evaluated tools. R denotes read and W denotes write. Names and parameters are shown as exposed. These abbreviated payloads can differ from the full operational API.}
\label{tab:full-api}
\end{table*}

\paragraph{Interface scope.}
The function schema is the model's direct contract. The full API-method count cannot substitute for exposed-tool count, and results on omitted operations require an access-limited interpretation.

\section{Seed Corpus Details}
\label{app:corpus-details}

The corpus is anchored at reference time \textbf{2001-10-15T10:00:00Z} (Monday morning).  All temporal offsets (\texttt{hoursAgo}, \texttt{daysAgo}, \texttt{daysFromNow}) are computed deterministically relative to this anchor.

\subsection{Persona}

The primary user is \textbf{Vince Kaminski} (\texttt{vkamins@enron.com}), VP \& Head of Research at Enron Corp, located on Floor~19 of the Enron Tower.
The corpus includes 16 users total: 10 Enron executives/employees, 2 Dynegy external partners, 3 system/bot accounts, and Kaminski himself.

\begin{table*}[t]
\centering
\scriptsize
\setlength{\tabcolsep}{3pt}
\begin{tabularx}{\textwidth}{@{}p{2.4cm}p{4.3cm}p{2.1cm}>{\raggedright\arraybackslash}X@{}}
\toprule
\textbf{Name} & \textbf{Email} & \textbf{Dept.} & \textbf{Title} \\
\midrule
Jeff Skilling     & jeff.skilling@enron.com  & Executive & President \& COO \\
Louise Kitchen    & louise.kitchen@enron.com & Trading   & Pres., EnronOnline \\
Sally Beck        & sally.beck@enron.com     & Operations & VP Operations \\
John Lavorato     & john.lavorato@enron.com  & Trading   & CEO, Enron Americas \\
Richard Shapiro   & richard.shapiro@enron.com & Gov. Affairs & VP Gov. Affairs \\
Steven Kean       & steven.kean@enron.com    & Executive & VP \& Chief of Staff \\
Mark Taylor       & mark.taylor@enron.com    & Legal     & VP \& Gen. Counsel \\
Kenneth Lay       & kenneth.lay@enron.com    & Executive & Chairman \& CEO \\
Stinson Gibner    & stinson.gibner@enron.com & Research  & Senior Analyst \\
Zimin Lu          & zimin.lu@enron.com       & Research  & Quant. Analyst \\
\midrule
Tim Belden        & tim.belden@dynegy.com    & Trading   & VP Western Trading \\
Rebecca Mark      & rebecca.mark@dynegy.com  & Development & VP Bus. Dev. \\
\midrule
EnronOnline Sys.  & system@enrononline.com   & N/A       & (bot) \\
Legal Ticketing   & legal-tickets@enron.com  & N/A       & (bot) \\
GasBank Platform  & alerts@gasbank.enron.com & N/A       & (bot) \\
\bottomrule
\end{tabularx}
\caption{All 15 non-self users in the corpus, spanning executives, analysts, external partners, and system accounts.}
\label{tab:corpus-users}
\end{table*}

\subsection{Message Distribution}

The corpus defines 40 recent incoming messages, 15 sent messages, and 10 older messages. Actual \texttt{parentFolderId} values give 50 Inbox messages, 15 Sent Items messages, and none in Archive: older-message construction retains the Inbox default.

Folder metadata differ from those object counts (for example, Inbox reports 42 and Sent Items reports 120). The corpus and frozen-read changelog implementation are preserved here. These discrepancies limit count-based and mutation-followup interpretations.

\noindent Key conversation threads: \texttt{conv-raptor} (4 messages about SPE financing), \texttt{conv-weather} (2 messages on weather derivatives), \texttt{conv-ferc} (2 messages on FERC filings), \texttt{conv-montecarlo} (2 messages on simulation parameters), \texttt{conv-dynegy} (2 messages on joint venture).

\subsection{Calendar Events}

The 15 events span October 8--23, 2001, around the October 15 reference time:
\begin{itemize}[leftmargin=*,itemsep=1pt]
\item \textbf{Before the reference time:} Project Raptor Kickoff (Oct 8) and Q3 Trading Retrospective (Oct 12).
\item \textbf{Oct 15:} Research Team Standup, 1:1 with Skilling, Risk Committee Meeting, and Gas Volatility Sync.
\item \textbf{Oct 16:} FERC Compliance Review, Dynegy Partnership Review, and 1:1 with Kitchen.
\item \textbf{Oct 17--20:} Focus Time and Project Raptor Deep Dive (Oct 17), Team Lunch at Pappadeaux (Oct 18), Trading Division All-Hands (Oct 19), and Dynegy Credit Review (Oct 20).
\item \textbf{Oct 22--23:} Energy Risk Conference.
\end{itemize}

\noindent FERC Compliance Review and Dynegy Partnership Review overlap on October 16 from 11:00 to 12:00 UTC. The date helper calls \texttt{setUTCHours(hour,0,0,0)}, which truncates fractional hours. Consequently, \texttt{evt-001}, \texttt{evt-002}, \texttt{evt-005}, and \texttt{evt-010} have zero duration. Some duration and free-slot rubrics expect fractional-hour boundaries instead, so affected timing-task grades have fixture or rubric validity limitations. The reported results preserve the evaluated fixture and saved grades.

\subsection{Other Entities}

\begin{itemize}[leftmargin=*,itemsep=1pt]
\item \textbf{15 contacts}: 10 Enron executives, 2 Dynegy partners, 3 institutional (Enron Treasury, NYMEX Operations, FERC Filings)
\item \textbf{3 todo lists}: ``Tasks'' (4 tasks), ``Research Items'' (3 tasks), ``Project Raptor Backlog'' (3 tasks), 10 tasks total spanning VaR model review, forward curve updates, and SPE valuation
\item \textbf{2 boards}: ``Project Raptor'' (4 tasks across To Do/In Progress/Done) and ``Q4 Research Goals'' (3 tasks)
\item \textbf{6 mail folders}: Inbox, Sent Items, Drafts, Deleted Items, Archive, Junk Email
\item \textbf{1 mail filter}: Auto-move EnronOnline trade confirmations
\item \textbf{3 contact groups}: Research Team, Risk Committee, Regulatory Affairs
\item \textbf{11 attachments}: Spreadsheets, PDFs, and documents across 9 messages
\end{itemize}

\section{LLM-Judge Protocol}
\label{app:judge-protocol}

\subsection{Judge System Prompt}

The checked-in evaluator uses judge deployment \texttt{dev-gpt-5-chat-jj} and does not specify temperature. It receives the following system prompt:

\begin{lstlisting}[language={},basicstyle=\ttfamily\scriptsize,frame=single,breaklines=true,columns=flexible]
You are an expert evaluator for an AI email
assistant benchmark.
You will be given:
1. The user's original query
2. The agent's final answer
3. A list of tool calls the agent made
4. A rubric describing what to check

Score the agent's answer on a scale of 0.0
to 1.0:
- 1.0 = fully satisfies the rubric
- 0.5 = partially satisfies
- 0.0 = completely fails

Respond with ONLY a JSON object:
{"score": <number>, "reason": "<brief
explanation>"}
\end{lstlisting}

\subsection{Seven Semantic Dimensions}

Each LLM assertion targets one of seven semantic aspects:

\begin{enumerate}[leftmargin=*,itemsep=2pt]
\item \textbf{Correctness}: Is the answer factually accurate given the corpus data? (e.g., correct counts, names, dates)
\item \textbf{Completeness}: Does the answer include \emph{all} required information, or does it omit key details?
\item \textbf{Relevance}: Is the response pertinent to the query, without extraneous tangents?
\item \textbf{Format}: Is the output well-structured and clearly presented? (e.g., tables for tabular data, bullet points for lists)
\item \textbf{Safety}: Does the agent avoid harmful, unauthorized, or destructive actions beyond what was requested?
\item \textbf{Action-taken}: Did the agent perform the correct mutations? (Used for write scenarios where the rubric checks that specific API calls were made correctly.)
\item \textbf{No-hallucination}: Does the answer avoid fabricating data not present in the corpus? (e.g., inventing email subjects, contact details, or meeting times)
\end{enumerate}

\paragraph{Rubric Design.}
Each rubric is hand-written by the scenario author and specifies exact entities, IDs, and expected relationships.
For example, a rubric for OB-H300 (cross-domain correlation) explicitly names the expected intersection of senders and attendees: ``Key people: Skilling (6 emails, in Research Standup/1:1/Risk Committee), Beck (6 emails, in Risk Committee), Kitchen (5 emails, in many events)\ldots Must NOT include EnronOnline/GasBank system accounts.''

\subsection{Judge Input Format}

For each LLM assertion, the judge receives a user message containing:

\begin{lstlisting}[language={},basicstyle=\ttfamily\scriptsize,frame=single,breaklines=true,columns=flexible]
## Query
<scenario query>

## Agent's Answer
<final text answer from agent>

## Tool Calls Made
  tool_name(args...) -> result...
  tool_name(args...) -> result...
  (arguments: 100 chars; result: 200 chars)

## Rubric to evaluate
**<label>**: <rubric text>

Respond with ONLY:
  {"score": <0.0-1.0>, "reason": "..."}
\end{lstlisting}

The evaluated judge receives at most 100 characters of arguments and 200 of results per call. This incomplete trace can limit its ability to verify execution details.

\clearpage
\section{Example Agent Trajectories}
\label{app:trajectories}
These summaries use recorded Claude Sonnet 4.5 tool calls from the result file with timestamp 2026-05-04T00:10:56.125Z. Iterations below are one-based. Arguments and results are abbreviated. The files preserve final answers and grading details but not every intermediate assistant message or untruncated response.

\subsection{Successful Trace: OB-N520 (folders, D3)}
The query creates Projects with Raptor, Dynegy, and FERC subfolders and routes matching messages (Appendix~\ref{app:scenario-examples}).

\begin{table}[t]
\centering
\scriptsize
\setlength{\tabcolsep}{3pt}
\begin{tabularx}{\columnwidth}{@{}p{0.45cm}>{\raggedright\arraybackslash}X@{}}
\toprule
Iter & Recorded calls and results (summary) \\
\midrule
1 & \texttt{folders\_create}: Projects, returning folder-5001. \\
2 & Three \texttt{folders\_create} calls: Raptor, Dynegy, and FERC, all with parent folder-5001 and returning folder-5002/5003/5004. \\
3 & Three \texttt{messages\_search} calls, one per keyword. \\
4 & Three \texttt{messages\_batchMove} calls. Results contain 6, 8, and 8 succeeded IDs and empty failed lists. \\
5 & Final answer reports the hierarchy and move counts. \\
\bottomrule
\end{tabularx}
\caption{Recorded OB-N520 success: five iterations, ten calls, and 34.886 s. Static, LLM, and overall scores are all 1.0. The scenario passes.}
\label{tab:traj-success}
\end{table}
All five critical static checks and the critical LLM rubric pass. Empty batch \texttt{failed} lists are successful results, not API errors. The trace uses returned folder IDs in later calls.

\subsection{Incomplete Trace: OB-H307 (inbox-cleanup, D3)}
The query requests ordered automated/high-importance/unread triage while excluding Sent Items.

\begin{table}[t]
\centering
\scriptsize
\setlength{\tabcolsep}{3pt}
\begin{tabularx}{\columnwidth}{@{}p{0.65cm}>{\raggedright\arraybackslash}X@{}}
\toprule
Iter & Recorded calls and results (summary) \\
\midrule
1 & Creates Action Required, FYI Only, and Automated folders (folder-5001/5002/5003). \\
2--5 & Retrieves Sent Items, lists messages, searches three automated senders, lists high-importance messages, and attempts an unread filter. \\
6 & Moves three EnronOnline matches in a batch and two other automated messages individually. \\
7 & Moves two further automated messages and six high-importance messages. \\
8--11 & Lists messages, attempts a folder filter, retrieves four folders, and lists folders. \\
12 & Creates another set of the same three named folders. \\
13--15 & Continues inbox listing with folderId and skip arguments until the iteration limit. \\
\bottomrule
\end{tabularx}
\caption{Recorded OB-H307 failure: 15 iterations, 29 calls, and 88.868 s. All six critical static checks pass with a static score of 1.0. The critical LLM rubric fails with a score of 0.0, yielding an overall score of 0.50 and a failing outcome.}
\label{tab:traj-fail}
\end{table}
The final answer is \emph{Let me continue getting all inbox messages:}. The judgment records incomplete execution and a missing final count report. The critical LLM failure prevents passing despite reaching the score threshold.

The trace also exposes environment and grading limits: message collection reads remain frozen, folder counts are not a live mutation view, and the static checks do not establish complete ordered triage. Some folder lookups do consult recorded creations. The recorded result supports an incomplete-workflow example.

\section{Per-Difficulty Breakdown (All Models)}
\label{app:difficulty-full}

Table~\ref{tab:difficulty-full} extends Table~\ref{tab:difficulty-results} with all eight configurations and additional metrics.

\begin{table*}[t]
\centering
\small
\setlength{\tabcolsep}{3pt}
\begin{tabular}{@{}lrrrrrrrrr@{}}
\toprule
Level & $n$ & Son. & Opus & 5.4-R & 5.5-R & 5.4 & 5.2 & o3 & GPT-5 \\
\midrule
D1 & 17 & 11.8 & 11.8 & 5.9 & 11.8 & 11.8 & 5.9 & 11.8 & 11.8 \\
D2 & 121 & 42.1 & 38.8 & 37.2 & 33.1 & 31.4 & 30.6 & 30.6 & 28.1 \\
D3 & 68 & 23.5 & 17.6 & 17.6 & 22.1 & 19.1 & 13.2 & 10.3 & 10.3 \\
\bottomrule
\end{tabular}
\caption{Pass rates (percent) for all eight configurations by assigned difficulty. These groups differ in tasks and tool availability. Differences do not isolate difficulty as a causal factor.}
\label{tab:difficulty-full}
\end{table*}

\paragraph{Interpreting assigned difficulty.}
D1 labels do not guarantee tool availability or easy grading. Some D1 tasks request board operations that the adapter cannot expose. The observed D1/D2/D3 differences therefore cannot be explained solely by model knowledge of rare endpoints.

\section{Scenarios Failed by the Four-Configuration Cohort}
\label{app:hard-scenarios}

117 scenarios (56.8\% of 206) are passed by none of the named Sonnet/Opus/GPT-5.4-R/o3 cohort. Table~\ref{tab:hard-profile} compares their declared metadata with the full suite.

\begin{table}[t]
\centering
\scriptsize
\setlength{\tabcolsep}{3pt}
\begin{tabular}{@{}lrr@{}}
\toprule
Attribute & Failed by four (117) & All (206) \\
\midrule
Mean assigned difficulty & 2.27 & 2.25 \\
Mean declared domains & 1.73 & 1.59 \\
Mean static assertions & 1.57 & 1.25 \\
Mean LLM assertions & 1.01 & 1.02 \\
D3 fraction & 40.2\% & 33.0\% \\
Calendar domain & 45.3\% & 34.5\% \\
Boards domain & 11.1\% & 9.2\% \\
Two or more domains & 46.2\% & 40.8\% \\
\bottomrule
\end{tabular}
\caption{Scenario metadata for tasks failed by Sonnet, Opus, GPT-5.4-R, and o3. Attributes overlap. This selected four-configuration cohort is not the top four by pass rate.}
\label{tab:hard-profile}
\end{table}

This cohort includes tasks with unavailable operations, persona mismatches, and grading limitations, as well as potentially difficult model behaviors. Its composition cannot be interpreted as a pure ranking of intrinsic reasoning difficulty.

\section{Iteration Distribution per Model}
\label{app:iter-dist}

Table~\ref{tab:iter-dist-full} shows the complete iteration count distribution and summary statistics for each configuration.

\begin{table}[t]
\centering
\scriptsize
\setlength{\tabcolsep}{3pt}
\begin{tabular}{@{}lrrrrrrr@{}}
\toprule
Model & $\mu$ & Med & $\sigma$ & Max & 1 & $\geq10$ & $\geq15$ \\
\midrule
Sonnet 4.5 & 3.66 & 3 & 2.67 & 15 & 7 & 9 & 6 \\
Opus 4.1 & 5.52 & 4 & 3.91 & 15 & 4 & 33 & 15 \\
GPT-5.4-R & 4.58 & 3 & 3.19 & 15 & 11 & 14 & 7 \\
GPT-5.5-R & 4.63 & 4 & 2.60 & 13 & 6 & 10 & 0 \\
GPT-5.4 & 3.09 & 2 & 1.88 & 12 & 13 & 4 & 0 \\
GPT-5.2 & 2.33 & 2 & 1.77 & 12 & 72 & 4 & 0 \\
o3 & 7.31 & 5 & 5.10 & 15 & 25 & 66 & 44 \\
GPT-5 & 2.35 & 2 & 1.52 & 11 & 71 & 1 & 0 \\
\bottomrule
\end{tabular}
\caption{Observed iteration distribution. $\sigma$ is the population standard deviation. Counts in the last three columns use all 206 scenarios. Missing stored counts are replaced by observed lower bounds from call indices.}
\label{tab:iter-dist-full}
\end{table}

\paragraph{Interpretation.}
Iteration counts include tool-calling steps and terminal answer steps when recorded. A one-iteration trajectory need not be a refusal, and a high count need not be an error-retry loop. Four exception records contain calls despite a zero stored iteration count. Their reconstructed counts are lower bounds.

\section{Reproducibility Details}
\label{app:reproducibility}

\paragraph{Recorded deployments and request settings.}
Table~\ref{tab:run-identifiers} gives the internal endpoint IDs, run IDs, and timestamps preserved in the eight saved baseline files. These service routing identifiers do not establish corresponding public model snapshots. All eight agent configurations and the judge use hosted deployments through the same Substrate bridge, including the Anthropic-labeled configurations. The checked-in agent and judge requests omit temperature and specify \texttt{max\_completion\_tokens=4096} and 2048, respectively. Agent and judge bridge subprocess limits are 180 and 120 seconds per request, and the underlying HTTP timeout is 120 seconds. One saved o3-labeled trajectory records 243.575 seconds of agent latency, confirming that the agent limit is not a scenario-wide deadline. Recorded latency excludes subsequent grading calls.

\paragraph{Runtime and cost provenance.}
The saved artifacts do not establish exact historical workstation or runtime versions, effective hosted endpoint defaults, complete-suite wall-clock duration, or billed token usage. We therefore make no fixed hardware, runtime-version, or cost claim.

\paragraph{Evaluation Variance.}
The analysis includes one recorded run per configuration. Repeated-run variance and expert-annotation agreement are not available.
\paragraph{Artifact provenance.}
Files include model IDs, run IDs, timestamps, final answers, tool traces, and grades. They omit complete source-commit and tool-schema manifests, and tool responses may be truncated. Offline checks reproduce component scores, combined scores, and critical-check pass flags for all 1,648 saved results, but these omissions still limit exact reconstruction of the evaluated environment.

\paragraph{Software Dependencies.}
The local benchmark uses Node.js, TypeScript, and the repository's npm dependencies. Model and judge execution additionally requires Python with \texttt{requests}, Azure identity and Key Vault libraries, authorized credentials, and access to the configured Substrate service. Recomputing saved-result tables and figures is offline and does not call a model or judge.

\end{document}

%% file: figures/out/table_execution_pipeline.tex
\begin{tabular}{@{}lrrrr@{}}
\toprule
Model & Engage & Clean & Answer & Pass \\
\midrule
Sonnet 4.5 & 96.6 & 94.7 & 100.0 & 33.5 \\
Opus 4.1 & 98.1 & 96.6 & 100.0 & 29.6 \\
GPT-5.4-R & 95.1 & 88.8 & 97.1 & 28.2 \\
GPT-5.5-R & 97.1 & 88.3 & 99.0 & 27.7 \\
GPT-5.4 & 93.7 & 89.3 & 100.0 & 25.7 \\
o3 & 87.9 & 79.1 & 78.6 & 22.3 \\
GPT-5.2 & 65.0 & 58.7 & 100.0 & 22.8 \\
GPT-5 & 65.5 & 57.8 & 100.0 & 20.9 \\
\bottomrule
\end{tabular}

%% file: figures/out/table_tool_success.tex
\begin{tabular}{@{}lrrrr@{}}
\toprule
Model & Calls & Success\,\% & First\,\% & Failures \\
\midrule
Sonnet 4.5 & 1170 & 99.7 & 100.0 & 4 \\
Opus 4.1 & 1107 & 99.7 & 100.0 & 3 \\
GPT-5.4-R & 743 & 96.8 & 100.0 & 24 \\
GPT-5.5-R & 1782 & 94.1 & 100.0 & 106 \\
GPT-5.4 & 547 & 95.4 & 100.0 & 25 \\
o3 & 1342 & 95.8 & 100.0 & 57 \\
GPT-5.2 & 383 & 92.4 & 100.0 & 29 \\
GPT-5 & 378 & 92.1 & 100.0 & 30 \\
\bottomrule
\end{tabular}

%% file: figures/out/table_recovery.tex
\begin{tabular}{@{}lrrrr@{}}
\toprule
Model & Failures & Next calls & Success\,\% & Same\,\% \\
\midrule
Sonnet 4.5 & 4 & 1 & 100.0 & 0.0 \\
Opus 4.1 & 3 & 0 & N/A & N/A \\
GPT-5.4-R & 24 & 17 & 52.9 & 47.1 \\
GPT-5.5-R & 106 & 98 & 14.3 & 86.7 \\
GPT-5.4 & 25 & 17 & 5.9 & 94.1 \\
o3 & 57 & 46 & 30.4 & 71.7 \\
GPT-5.2 & 29 & 18 & 33.3 & 66.7 \\
GPT-5 & 30 & 14 & 0.0 & 100.0 \\
\bottomrule
\end{tabular}